\documentclass[lettersize,journal]{IEEEtran}
\usepackage{amsfonts}
\usepackage{array}
\usepackage[caption=false,font=normalsize,labelfont=sf,textfont=sf]{subfig}
\usepackage{textcomp}
\usepackage{stfloats}
\usepackage{url}
\usepackage{verbatim}
\usepackage{graphicx}
\usepackage{cite}
\usepackage{mathtools}
\usepackage{amsthm}
\usepackage{amsmath}
\usepackage{amssymb}
\usepackage{algpseudocode}
\usepackage[colorlinks,bookmarksopen,bookmarksnumbered,citecolor=blue, linkcolor=blue, urlcolor=black]{hyperref}
\theoremstyle{plain}
\newtheorem{theorem}{Theorem}

\theoremstyle{definition}

\newtheorem*{lemma*}{Lemma}
\newtheorem{corollary*}{Corollary}
\usepackage[ruled,vlined,noend]{algorithm2e}
\usepackage{wrapfig}
\usepackage{multirow}
\usepackage{tablefootnote}
\usepackage{subcaption}
\usepackage[normalem]{ulem}
\useunder{\uline}{\ul}{}

\def\eqref#1{(\ref{#1})}
\def\cS{{\mathcal{S}}}

\def\cD{{\mathcal{D}}}
\def\cA{{\mathcal{A}}}
\def\cN{{\mathcal{N}}}

\def\cL{{\mathcal{L}}}

\def\cZ{{\mathcal{Z}}}
\def\cW{{\mathcal{W}}}
\def\cH{{\mathcal{H}}}

\def\EE{{\mathbb{E}}}

\def\cP{{\mathcal{P}}}
\def\RR{{\mathbb{R}}}

\begin{document}
 
\title{Unsupervised Skill Discovery through Skill Regions Differentiation}

\author{
\IEEEauthorblockN{
Ting Xiao\textsuperscript{1}\IEEEauthorrefmark{1},
Jiakun Zheng\textsuperscript{1},
Rushuai Yang\textsuperscript{3},
Kang Xu\textsuperscript{4},
Qiaosheng Zhang\textsuperscript{5},
Peng Liu\textsuperscript{6},
Chenjia Bai\textsuperscript{2}
}\\
\IEEEauthorblockA{
\textsuperscript{1}East China University of Science and Technology\\
}
\IEEEauthorblockA{
\textsuperscript{2}Institute of Artificial Intelligence (TeleAI), China Telecom\\
}
\IEEEauthorblockA{
\textsuperscript{3}Hong Kong University of Science and Technology
}
\IEEEauthorblockA{
\textsuperscript{4}Tencent\\
}
\IEEEauthorblockA{
\textsuperscript{5}Shanghai AI Laboratory
}
\IEEEauthorblockA{
\textsuperscript{6}Harbin Institute of Technology\\
}
\IEEEauthorrefmark{1}Corresponding author.
}

% The paper headers
\markboth{Journal of \LaTeX\ Class Files,~Vol.~14, No.~8, August~2021}%
{Shell \MakeLowercase{\textit{et al.}}: A Sample Article Using IEEEtran.cls for IEEE Journals}

% Remember, if you use this you must call \IEEEpubidadjcol in the second
% column for its text to clear the IEEEpubid mark.

\maketitle

\begin{abstract}
Unsupervised Reinforcement Learning (RL) aims to discover diverse behaviors that can accelerate the learning of downstream tasks. Previous methods typically focus on entropy-based exploration or empowerment-driven skill learning. However, entropy-based exploration struggles in large-scale state spaces (e.g., images), and empowerment-based methods with Mutual Information (MI) estimations have limitations in state exploration. To address these challenges, we propose a novel skill discovery objective that maximizes the deviation of the state density of one skill from the explored regions of other skills, encouraging inter-skill state diversity similar to the initial MI objective. For state-density estimation, we construct a novel conditional autoencoder with soft modularization for different skill policies in high-dimensional space. Meanwhile, to incentivize intra-skill exploration, we formulate an intrinsic reward based on the learned autoencoder that resembles count-based exploration in a compact latent space. Through extensive experiments in challenging state and image-based tasks, we find our method learns meaningful skills and achieves superior performance in various downstream tasks.
\end{abstract}

\begin{IEEEkeywords}
 Unsupervised reinforcement learning, skill discovery, inter-skill diversity, intra-skill exploration.
\end{IEEEkeywords}

\section{Introduction}
\IEEEPARstart{R}{einforcement} Learning (RL) has achieved remarkable success in game AI \cite{alphazero, efficientZero}, autonomous cars \cite{driving-nature, driving}, and embodied agents \cite{TD-MPC,quad-science-2022}. 
% However, these successes mainly rely on well-defined reward functions based on domain knowledge \cite{kwon2023reward,ManiSkill2}, limiting the generalization of learned policies to other tasks with different objectives. 
Traditionally, RL agents rely on well-designed reward functions to learn specific tasks \cite{luo2023relay}. However, designing these reward functions is resource-intensive and often requires domain-specific expertise \cite{kwon2023reward,ManiSkill2}, making the learned policies dependent on handcrafted rewards and potentially unable to capture the complexity of real-world scenarios. This reliance limits the agent’s generalization capability across diverse tasks and results in poor adaptability.
In contrast, recent advances in Large Language Models (LLMs) \cite{NLP-1,GPT4} signify that unsupervised auto-regression has led to powerful pre-trained language models, which can be adapted to downstream tasks via supervised fine-tuning \cite{InstructGPT,llama}. A powerful vision encoder can also be pre-trained via masked prediction 
% or contrastive learning 
without annotations or labels \cite{mae,vjepa,BYOL}, and the encoder can be used to solve various vision tasks \cite{VC1,R3M}. Inspired by these breakthroughs, it is desirable to further explore similar unsupervised learning methods within the RL field. The goal is for unsupervised RL to learn useful behaviors in the absence of external rewards, thus equipping them with the capacity to quickly adapt to new tasks with limited interactions \cite{URLB}.
% it is also desirable for an RL agent to learn useful behaviors without extrinsic rewards, allowing the behaviors to quickly adapt to downstream RL tasks with a small number of interactions \cite{URLB}. 

The formulation of unsupervised RL has been studied in many prior works, which can be roughly categorized into empowerment-based skill discovery \cite{vic} and pure exploration methods \cite{apt}. Empowerment-based methods aim to maximize the Mutual Information (MI) between states and skills, and the MI term can be estimated by different variational estimators \cite{MI-estimator}. These methods have shown effectiveness in learning discriminative skills for state-based locomotion tasks \cite{diayn}. However, the learned skills often have limited state coverage due to the inherent sub-optimality in the MI objective \cite{becl}, which can lead to sub-optimal adaptation performance in downstream tasks and becomes more severe in large-scale state space \cite{park2024metra}. Recent works introduce additional techniques like Lipschitz constraints and metric-aware abstraction to enhance the exploration abilities \cite{LSD-2022,USD-2023,park2024metra}. Pure exploration methods encourage the agent to explore the environment with maximum state coverage; however, this can lead to extremely dynamic skills rather than meaningful behaviors for downstream tasks \cite{apt,cic}. Meanwhile, both the MI estimator and entropy estimation are not directly scalable to large-scale spaces, such as pixel-based environments \cite{pixel-urlb,park2024metra}. 

To overcome the aforementioned limitations, this work proposes a novel skill discovery method by maximizing the \emph{State Density Deviation of Different skills} (\textbf{SD3}). Specifically, we construct a conditional autoencoder for state density estimation of different skills in high-dimensional state spaces. Each skill policy is then encouraged to explore regions that deviate significantly from the state density of other skills, which encourages \emph{inter-skill diversity} and leads to discriminative skills. For a stable state-density estimation of significantly different skills, we adopt soft modularization for the conditional autoencoder to make the skill-conditional network a weighted combination of the shared modules according to a routing network determined by the skill. We show the skill-deviation objective of SD3 resembles the initial MI objective in a special case. Further, to incentivize \emph{intra-skill exploration}, we formulate an intrinsic reward from the autoencoder based on the learned latent space, which extracts the skill-relevant information and is scalable to large-scale problems. Theoretically, such an intrinsic reward is closely related to the provably efficient count-based exploration in tabular cases. To summarize, SD3 encourages inter-skill diversity via density deviation and intra-skill exploration via count-based exploration in a unified framework. We conduct extensive experiments in Maze, state-based Unsupervised Reinforcement Benchmark (URLB), and challenging image-based URLB environments, showing that SD3 learns exploratory and diverse skills. 

Our contribution can be summarized as follows. (\romannumeral1) We propose a novel skill discovery objective based on state density deviation of skills, providing a straightforward way to learn diverse skills with different state occupancy. (\romannumeral2) We propose a novel conditional autoencoder with soft modularization to estimate the state density of significantly different skills stably. (\romannumeral3) The learned latent space of the autoencoder provides an intrinsic reward to encourage intra-skill exploration that resembles count-based exploration in tabular MDPs. (\romannumeral4) Our method achieves state-of-the-art performance in various downstream tasks in challenging URLB benchmarks and demonstrates scalability in image-based URLB tasks. 

\section{Preliminaries}
\subsection{Markov Decision Process}
A Markov Decision Process (MDP) constitutes a foundational model in decision-making scenarios. We consider the process of an agent interacting with the environment as an MDP with discrete skills, defined by a tuple $(\cS,\cA,\cZ,\cP,r,\gamma)$, where $\cS$ is the state space, $\cA$ is the action space, $\cZ$ is the skill space, $\cP:\cS\times\cA\to\Delta(\cS)$ is the transition function, $r:\cS\times\cA\to\mathbb{R}$ is the reward function, and $\gamma$ is the discount factor. In this work, we consider a discrete skill space $\cZ$ that contains $n$ skills since calculating the skill density deviation requires density estimation of all skills, while SD3 can also be extended to a continuous skill space by sampling skills from a continuous distribution for approximation. 
% We use $z\in\RR^n$ to denote a one-hot skill vector $(0,...,1,...,0)$ where the $i$-th coordinate is set to 1 and other coordinates are set to 0.
In each timestep, an agent follows a skill-conditional policy $\pi(a|s,z)$ to interact with the environment. Given clear contexts, we refer to `skill-conditional policy' as `skill'.

\subsection{Unsupervised RL}
Unsupervised RL typically contains two stages: unsupervised pre-training and fast policy adaptation. In the unsupervised training stage, the agent interacts with the environment without any extrinsic reward. The policy $\pi(a|s,z)$ is learned to maximize some intrinsic rewards $r_t$ formulated by an estimation of the MI term or the state entropy. The aim of unsupervised pre-training is to learn a set of useful skills that potentially solve various downstream tasks via fast policy adaptation. In the adaptation stage, the policy $\pi(a|s,z^\star)$ with a chosen skill $z^\star$ is optimized by RL algorithms with certain extrinsic rewards to adapt to specific downstream tasks. In the following, we denote $I(\cdot;\cdot)$ by the MI between two random variables and $\cH(\cdot)$ by either the Shannon entropy or differential entropy, depending on the context. We use uppercase letters for random variables and lowercase letters for their realizations. We denote $d^\pi(s)\triangleq (1-\gamma)\sum_{t=0}^\infty \gamma^t P(s_t=s|\pi)$ as the normalized probability that a policy $\pi $ encounters state $s$. 

The empowerment-based skill discovery algorithms try to estimate the MI between $S$ and $Z$ via $I(S;Z)=\mathbb{E}_{z\sim p(z),s\sim p^{\pi}(s|z)}[\log p(z|s)-\log p(z)]$. Given the computational challenges associated with the posterior $p(z|s)$, a learned skill discriminator $q_{\phi}(z|s)$ is employed \cite{diayn} and a variational lower bound is established for the MI term as $I(Z;S) \geq \mathbb{E}_{z\sim p(z),s\sim p^{\pi}(s|z)}[\log q_\phi(z|s)-\log p(z)]$. Alternatively, pure exploration methods estimate state entropy by summing the log-distances between each particle and its $k$-th nearest neighbor, as $\cH(s)\propto \sum_{s_i}\ln\|s_i-{\rm NN}_k(s_i)\|$.

\section{Method}
In this section, we first introduce the proposed SD3 algorithm that performs skill discovery by maximizing inter-skill diversity via state density estimation. Next, we present the formulation of intrinsic rewards for intra-skill exploration. Finally, we provide a qualitative analysis of SD3. 
\subsection{Skill Discovery via Density Deviation}
\label{sec:skill-discovery}

We develop our skill discovery strategy from a straightforward intuition: The explored region of each skill should deviate from other skills as far as possible. Formally, the optimizing objective for skill discovery, denoted as $I_{\rm SD3}$ and referred to as \emph{density deviation}, is defined by
% Formally, we use $d^\pi_{z}(s)$ to measure the state density of skill $z$, then the optimizing objective for skill discovery is defined by
\begin{equation}
\label{eq:sd3-1}
I_{\rm SD3}\triangleq\EE_{z\sim p(z),s\sim d^{\pi}_{z}(s)}\left[\log \frac{\lambda\: d^{\pi}_{z}(s)}{\lambda\: d^{\pi}_{z}(s) p(z) +\sum_{z'\neq z} d^{\pi}_{z'}(s) p(z')}\right],
\end{equation}
where $z$ is sampled from $p(z)$, $s$ is sampled from the state distribution induced by the skill policy $\pi(a|s,z)$, and $\lambda>0$ is a weight parameter. The numerator $d^\pi_{z}(\cdot)$ is the state density of skill $z$, and the denominator is the weighted average of the state density of $z$ and those of other skills $\{z'\}$. Since we uniformly sample skills from the skill set that contains $n$ skills, we have $p(z)=1/n$ for each skill $z$.
% and 
% \begin{equation}
% \label{eq:sd3-2}
% I_{\rm SD3}=\EE_{z\sim p(z),s\sim d^{\pi}_{z}(s)}\left[\log \frac{\lambda n\: d^{\pi}_{z}(s)}{\lambda\: d^{\pi}_{z}(s) +\sum_{z'\neq z} d^{\pi}_{z'}(s)}\right].
% \end{equation}
According to Eq.~\eqref{eq:sd3-1}, it is easy to check that $I_{\rm SD3}$ attains its maximum when $\sum_{z'\neq z} d^{\pi}_{z'}(s)\rightarrow 0$ for all $(s,z)$ such that $p(z)\cdot d_z^{\pi}(s) > 0$, and the maximum value is $\mathcal{H}(Z)$.
% $\max I_{\rm SD3}=\log n$. 
In this case, the state $s\sim d^\pi_{z}(\cdot)$ visited by skill $z$ has \emph{zero} visitation probability by other skills, which means the explored regions of all skills do not overlap, and the learned skills are fully distinguishable. However, enforcing such a strong objective to separate the overlapping explored areas of skills may lead to limited state coverage for each skill. In extreme cases, each skill might only visit a distinct state that other skills do not access. Although this leads to distinguishable skills, the overall state coverage becomes overly limited, making them undesirable for learning meaningful behaviors.

In SD3, we adopt two mechanisms for addressing this problem. (\romannumeral1)~A weight parameter $\lambda$ is used in the learning objective to regularize the gradients of $I_{\rm SD3}$ to other skills.
%To see this, we denote the state density of other skills $\{z'\}$ except for $z$ in Eq.~\eqref{eq:sd3-1} as 
%$\rho_{z^c} \triangleq \sum_{z' \ne z} d_{z'}^{\pi}(s)$,
%then the gradient of $I_{\rm SD3}$ to $\rho_{z^c}$ becomes 
%\begin{equation}
%\nabla_{\rho_{z^c}} I_{\rm SD3}=\EE_{z,s}[-1/(\lambda d^\pi_{z}(s)+\rho_{z^c})].
%\end{equation}
To see this, for each $(s,z)$, we denote the state density of other skills $\{z'\}$ except $z$ as $\rho_{z^c} \triangleq \sum_{z' \ne z} d_{z'}^{\pi}(s)$, then the gradient of $I_{\rm SD3}(s,z)$ to $\rho_{z^c}$ becomes
\begin{equation}
\label{eq:gradient}
    \nabla_{\rho_{z^c}} I_{\rm SD3}(s,z)= -1/(\lambda d^\pi_{z}(s)+\rho_{z^c}(s)),
\end{equation}
where $I_{\rm SD3}(s,z)$ is the density ratio for a specific $(s, z)$ and the proof is attached in Appendix~\ref{sec:proof_grad}.
Thus, for skill $z$, increasing $\lambda$ will weaken the gradient of SD3 in reducing the state densities of other skills, which prevents skill collapse in SD3. (\romannumeral2) We introduce explicit intra-skill exploration based on the latent space learned in estimating the skill density, which will be discussed in \S\ref{sec:exploration}. To maximize $I_{\rm SD3}$, we adopt a modified Conditional Variational Auto-Encoder (CVAE) to stably estimate the state density for skills, which we introduce as follows.

\textbf{CVAE for State Density Estimation.} In SD3, we adopt a lower bound of skill-conditional state density (i.e., $\log d^\pi_z(s)$) via stochastic gradient variational Bayes. We adopt CVAE with a latent representation $h$ to obtain a variational form as
\begin{equation}
\begin{aligned}
\label{eq:elbo}
\log &d^\pi_z(s) = \EE_{Q(h |s,z)}\log\left[P(s|z)\right] 
\\&= \EE_{Q(h |s,z)}\log\left[\frac{P(s,h |z)}{Q(h|s,z)}\right] + \EE_{Q(h|s,z)}\log\left[\frac{Q(h|s,z)}{P(h|s,z)}\right] 
\\&\geq \EE_{Q(h |s,z)}\log\left[\frac{P(s|h, z)P(h|z)}{Q(h|s,z)}\right]
\\&= \underbrace{\EE_{Q(h |s,z)}\log\left[P(s|h, z)\right] -D_{\rm KL}\left[Q(h|s,z)\|P(h|z)\right]}_{\cL^{\rm elbo}_z(s)},
\end{aligned}
\end{equation}
where the latent vector $h$ is sampled from a variational posterior distribution (i.e., $Q(h|s,z)$) conditioned on the state and skill, and the inequality holds by dropping off the non-negative second term, which is the definition of $D_{\rm KL}(Q(h|s,z)\|P(h|s,z))$. Meanwhile, we use $P(s,h|z)=P(h|z)P(s|h,z)$ to decompose the joint distribution.
% \begin{equation}
% \label{eq:elbo}
% \log d^\pi_z(s)\geq \EE_{Q(h|s,z)}\big[\log P(s|h,z)\big] - D_{\rm KL}\big[Q(h|s,z)\|P(h|z)\big]\triangleq \cL^{\rm elbo}_z(s).
% \end{equation}
According to Eq.~\eqref{eq:elbo}, maximizing the Evidence Lower-Bound (ELBO) $\cL^{\rm elbo}_z(s)$ can approximate the skill-conditioned state distribution, as $\log d^\pi_z(s)\approx \max_{Q} \cL^{\rm elbo}_z(s)$. To maximize $\cL^{\rm elbo}_z(s)$, we learn an encoder network $Q_\phi(h|s,z)$ to obtain the posterior of latent representation, where the posterior is represented by a diagonal Gaussian. Then, a latent vector $h$ is sampled from the posterior, and a decoder network $P_\psi(s|h,z)$ is used to reconstruct the state. The KL-divergence in $\cL^{\rm elbo}_z(s)$ regularizes the latent space via a prior distribution $P(h|z)$, which is set to a standard Gaussian. The whole objective is optimized via stochastic gradient ascent with a reparameterization trick \cite{vae,vae-intro}.
% for gradient propagating across the latent space. 
To calculate $I_{\rm SD3}$, we perform state density estimations for all skills via forward inference based on the learned encoder and decoder. In calculating $I_{\rm SD3}$, we adopt efficient parallelization to calculate $\cL^{\rm elbo}_{z}(s)$ for all skills $z \in \mathcal{Z}$ in one forward pass, which minimizes the run-time increase with the number of skills.

\begin{figure*}[htbp]
\begin{center}
\centerline{
\includegraphics[width=\textwidth]{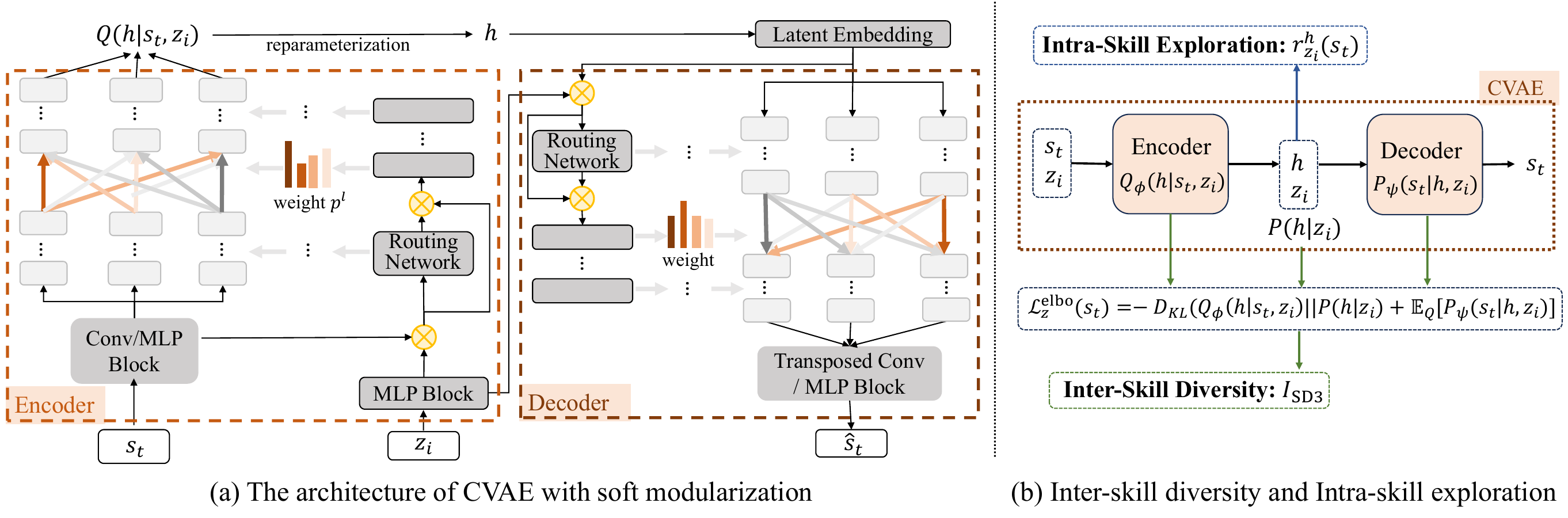}}
\caption{An overview of the CVAE architecture. (a) The encoder-decoder network with soft modularization. The feature extractor of state can be MLPs or convolution layers according to state- or image-based environment. (b) The inter-skill diversity objective for skill discovery and the intra-skill intrinsic reward for exploration can be derived from the learned CVAE.}
\label{fig:cvae_architecture}
\end{center}
\vspace{-2em}
\end{figure*}

\textbf{Soft Modularization for CVAE.} As we maximize the state-density deviation in skill discovery, the resulting skills become diverse, and the corresponding state occupancy for different skills tends to be very different. In CVAE-based density estimation, since different skills share the same network parameters, optimizing $\cL^{\rm elbo}_z$ for one skill can negatively affect the density estimation of other skills with significantly different state densities. 
Empirically, we also find obtaining an accurate estimation of $d^\pi_{z}(s)$ for all skills $z \in \mathcal{Z}$ can be difficult. As a result, we 
% use slightly different network modules for the density estimation of different skills. Specifically, we 
adopt a soft modularization technique that automatically generates soft network module combinations for different skills without explicitly specifying structures. As shown in Fig.~\ref{fig:cvae_architecture}, the soft modularized CVAE contains an unconditional basic network and a routing network, where the routing network takes the skill and state embedding as input to estimate the routing strategy. Suppose each layer of the encoder/decoder network has $m$ modules, then the routing network gives the probabilities $p\in\RR^{m\times m}$ to weight modules contributing to the next layer. Specifically, considering $l$-th layer has probabilities $p^l\in\RR^{m\times m}$, then the probability in the next layer is
\begin{equation}
\label{eq:prob_vector}
p^{l+1}=\cW^l\big({\rm ReLU}(g(p^l)\odot (u \odot v))\big), u=f_1(s),v=f_2(z), 
\end{equation}
where $\odot$ denotes element-wise product, $g(\cdot)$, $f_1(\cdot)$ and $f_2(\cdot)$ are all fully connected layers that $f_1(\cdot)$ and $f_2(\cdot)$ map state $s$ and skill $z$ to the same dimensions (e.g., $d$), and $g(\cdot)$ maps $p^l$ to the dimension $d$. Then we have $\cW^l\in\RR^{m^2\times d}$ to project the joint feature to a probability vector of layer $l+1$. In the basic network, we denote the input feature for the $j$-th module in the $l$-the layer as $g^{l}_j\in\RR^{d}$; then we have $g_i^{l+1}=\sum_{j}\hat{p}^l_{i,j}({\rm ReLU}(\cW_j^lg_j^l))$ for the next layer, where $\hat{p}^l_{i,j}=\exp(
p^l_{i,j})/(\sum_{j=1}^m\exp (p^l_{i,j}))$ is the normalized vector that weights the $j$-th module in the $l$-th layer to contribute to the $i$-th module in the $l+1$-th layer. We remark that the soft modularization technique was originally proposed in multi-task RL \cite{soft-module}, while we extend it to encoder-decoder-based CVAE for density estimation. 

\begin{figure}[!t]
\centering
\includegraphics[width=0.4\textwidth]{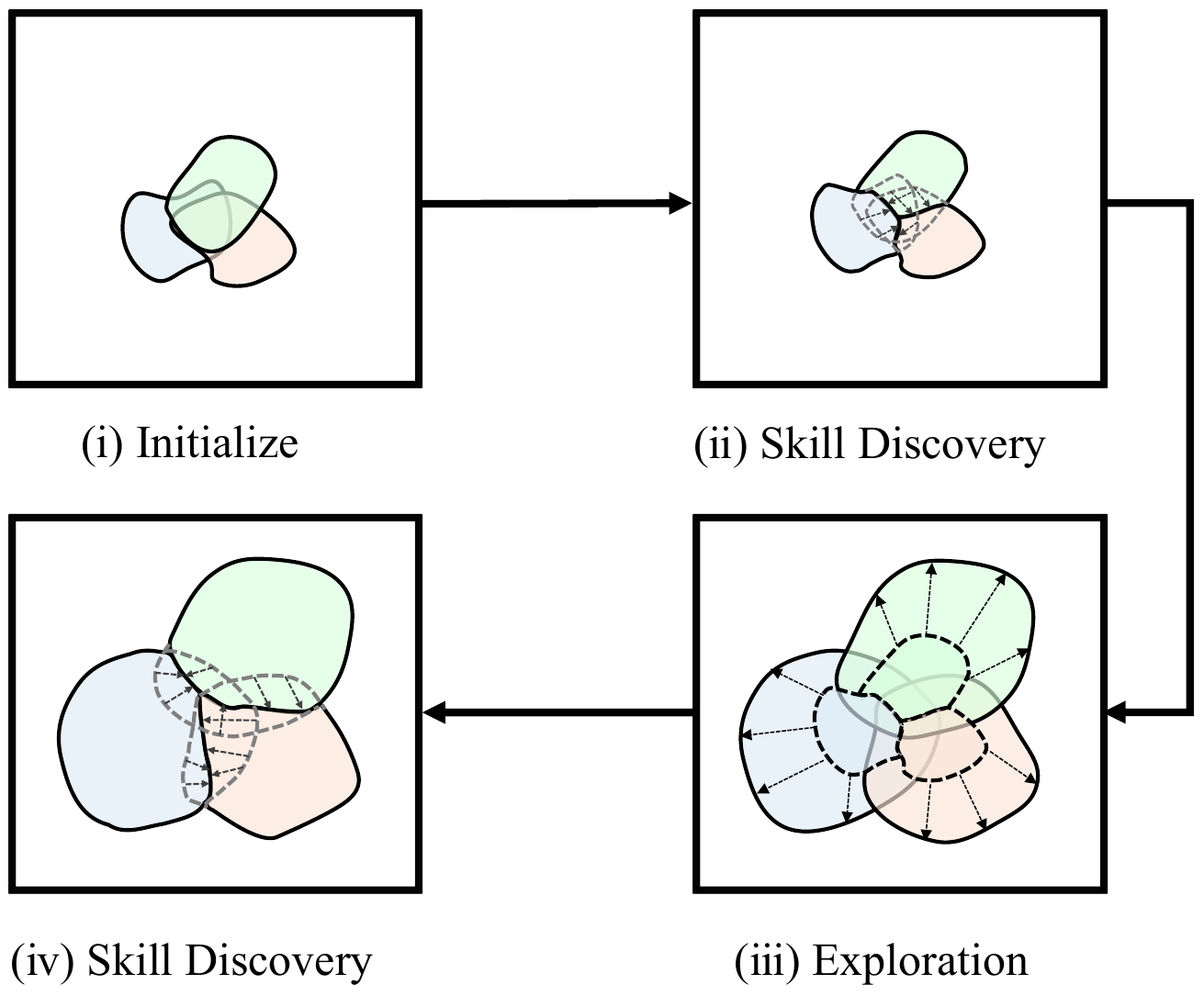}
\caption{An illustration of skill discovery in SD3. The skills start with overlapping areas and are separated via state-density deviation. Then, each skill explores the environment independently, resulting in overlapped but expanded areas. SD3 separates the areas again and leads to distinguishable skills. Such a process repeats and ultimately leads to exploratory and diverse skills.}
\label{fig:sd3-model}
\vspace{-1em}
\end{figure}

% \begin{figure*}[htbp]
%   \centering
%   \begin{minipage}{0.45\textwidth} 
%     \centering
%     \includegraphics[width=\linewidth]{Fig/sd3.pdf}
%   \end{minipage}%
%   \hfill
%   \begin{minipage}{0.5\textwidth}
%     \centering
%     \begin{algorithm}[H]
%       % \small
%       \DontPrintSemicolon
%       \SetAlgoLined
%         Initialize CVAE $Q_\phi$, $P_\psi$, policy $\pi_{\theta}$, $Q$-func, $Q_\varphi$, buffer $\cD$;\\
%         \While{not converged}{
%             Randomly choose $z$ from the skill set $\cZ$\;\\
%             Interact with environment with $\pi_\theta(a|s,z)$\;\\
%             Store the transitions in a replay buffer $\cD$\;\\
%             Sample a batch of transitions $\{(s_t,a_t,s'_t,z)\}\sim\cD$\;\\
%             Calculate $r_{z}^{\rm sd3}$ and $r_{z}^{\rm exp}$ for transitions via Eqs.~\eqref{eq:rew-exp}-\eqref{eq:rew-sd3}\;\\ 
%             Update CVAE parameters via $\cL^{\rm elbo}$ in Eq.~\eqref{eq:elbo} \;\\
%             Update policy and $Q$-value via off-policy algorithm \;
%         }
%       \caption{SD3 Algorithm}
%       \label{alg}
%     \end{algorithm}
%   \end{minipage}
%     \caption{An illustration of skill discovery in SD3. The skills start with overlapping areas and are separated via state-density deviation. Then, each skill explores the environment independently, resulting in overlapped but expanded areas. SD3 separates the areas again and leads to distinguishable skills. Such a process repeats and ultimately leads to exploratory and diverse skills.
% \label{fig:sd3-model}}
% \vspace{-20pt}
% \end{figure*}

\subsection{Latent Space Exploration}
\label{sec:exploration}

As we discussed above, the SD3 objective that only maximizes the density deviation may lead to skill collapse. In addition to introducing an additional parameter $\lambda$ in Eq.~\eqref{eq:sd3-1}, we find the learned CVAE in Fig.~\ref{fig:cvae_architecture} can provide a \emph{free-lunch} intrinsic reward for efficient intra-skill exploration. In SD3, we derive an intrinsic reward based on the latent space that learns skill-conditioned representations for states. Specifically, the KL-divergence term $D_{\rm KL}\big[Q(h|s,z)\|r(h)\big]$ in CVAE objective serves as an upper bound of the conditional MI term $I(S;H|Z)$, as
\begin{equation}
\begin{aligned}
\label{eq:MI-compress}
I(S;H|Z)&=\EE_{p(s,z),Q_\phi(h|s,z)}\big[\log Q_\phi(h|s,z) / P(h|z)\big]
\\&\leq \EE_{p(s,z),Q_\phi(h|s,z)}\big[\log Q_\phi(h|s,z) / r(h)\big],
\end{aligned}
\end{equation}
where $H$ denotes the random variable of the sampled latent representation $h$, and $r(h)$ the prior distribution set to a standard Gaussian, and $P(h|z) \triangleq \EE_{P(s|z)} Q_{\phi}(h|s,z)$. The inequality holds since $D_{\rm KL}[P(h|z)\|r(h)]\geq 0$ for all $z \in \cZ$. Since $D_{\rm KL}[Q_\phi(h|s,z) \| r(h)]$ is constrained in CVAE learning, the MI between states and latent representations for each skill is also compressed according to Eq.~\eqref{eq:MI-compress}. Thus, the latent space in CVAE learns a compressive representation while retaining important information as the representation is then used for reconstruction. Based on the learned representation, we define the intrinsic reward for intra-skill exploration as
\begin{equation}
\label{eq:rew-exp}
r^{\rm exp}_{z}(s) = D_{\rm KL}[Q_\phi(h|s,z)\|r(h)],
\end{equation}
where $Q_\phi(h|s,z)$ is the posterior network learned in CVAE. The intrinsic reward in Eq.~\eqref{eq:rew-exp} quantifies the degree of compression of representation with respect to the state, which measures skill-conditioned state novelty in a compact space for intra-skill exploration. Intuitively, if a state $s^{(1)}$ is frequently visited by skill $z$, then the corresponding latent distribution is close to $r(h)$ according to Eq.~\eqref{eq:MI-compress}, and the resulting reward $r^{\rm exp}_{z}(s^{(1)})$ will be close to zero. In contrast, if a state $s^{(2)}$ is novel for skill $z$, then the corresponding intrinsic reward will be high since the latent posterior $Q_\phi(h|s^{(2)},z)$ can be very different from the prior $r(h)$. Thus, in exploration, such reward encourages the policy to find the scarcely visited states
\begin{algorithm}[ht]
\caption{SD3 Algorithm}
\label{alg}
\DontPrintSemicolon
\SetAlgoLined
Initialize CVAE $Q_\phi$, $P_\psi$, policy $\pi_{\theta}$, $Q$-func, $Q_\varphi$, buffer $\cD$\;
\While{not converged}{
    Randomly choose $z$ from the skill set $\cZ$\;
    Interact with environment with $\pi_\theta(a|s,z)$\;
    Store the transitions in a replay buffer $\cD$\;
    Sample a batch of transitions $\{(s_t,a_t,s'_t,z)\}\sim\cD$\;
    Calculate $r_{z}^{\rm sd3}$ and $r_{z}^{\rm exp}$ for transitions via Eqs.~\eqref{eq:rew-exp}-\eqref{eq:rew-sd3}\;
    Update CVAE parameters via $\cL^{\rm elbo}$ in Eq.~\eqref{eq:elbo}\;
    Update policy and $Q$-value via off-policy algorithm\;
}
\end{algorithm}
$\{s^{+}\}$ (with a high $D_{\rm KL}[Q_\phi(h|s,z)\|r(h)]$) and explore these states.

An illustration of the skill learning process of SD3 is shown in Fig.~\ref{fig:sd3-model}. The state occupancy of different skills overlaps initially in Fig.~\ref{fig:sd3-model}(\romannumeral1), then we maximize $I_{\rm SD3}$ via per-instance estimation and set it to an intrinsic reward as
\begin{equation}
\label{eq:rew-sd3}
r^{\rm sd3}_{z}(s)=\log \frac{\lambda\: d^{\pi}_{z}(s)}{\lambda\: d^{\pi}_{z}(s) p(z) +\sum_{z'\neq z} d^{\pi}_{z'}(s) p(z')},
\end{equation}
which encourages skill density deviation and leads to more diverse skills with separate state coverage, as in Fig.~\ref{fig:sd3-model}(\romannumeral2). Then the exploration reward $r^{\rm exp}_{z}(s)$ is used to encourage intra-skill exploration, which makes each skill explore unknown areas independently. After exploration, the state coverage of each skill increases and may lead to state-coverage overlapping again among skills, as in Fig.~\ref{fig:sd3-model}(\romannumeral3). Then the density-derivation reward $r^{\rm sd3}_{z}(s)$ re-separates the updated areas to obtain distinguished skills, as in Fig.~\ref{fig:sd3-model}(\romannumeral4). The above process repeats for many rounds and SD3 finally learns exploratory and diverse skills. The algorithmic description of our method is given in Algorithm~\ref{alg}. 

\subsection{Qualitative Analysis}
\label{sec:qual_analysis}

In this section, we give a qualitative analysis of the proposed SD3 objective and exploration reward, which encourage inter-skill diversity and intra-skill exploration, respectively. 

The skill discovery objective $I_{\rm SD3}$ in Eq.~\eqref{eq:sd3-1} leads to diverse skills with separate explored areas, which is similar to the MI-based skill discovery objectives. As we usually set $\lambda\geq 1$ to prevent skill collapse, the following theorem connects $I_{\rm SD3}$ and the previous MI objectives. 
\begin{theorem}\label{thm:1}
With $\lambda \geq 1$, we have 
\begin{equation}
I(S;Z) \leq I_{\rm SD3} \leq c_0 + I(S;Z).
\end{equation}
% where $c_0=\log \lambda + \frac{1}{\lambda} - 1 + \frac{\lambda - 1}{ \lambda^2} \mathcal{H}(Z)$. 
where $c_0=\log \lambda$.
Specially, $I_{\rm SD3}=I(S;Z)$ if $\lambda=1$.
\end{theorem}
The above theorem shows when we maximize skill deviation via $I_{\rm SD3}$, the MI between $S$ and $Z$ also increases. The previous MI objective becomes a special case of $I_{\rm SD3}$, where the introduced $\lambda$ provides flexibility to control the strength of skill deviation. The proof of Theorem~\ref{thm:1} is attached in Appendix~\ref{sec:proof_thm1}. In the following, we connect the proposed intrinsic reward to the provably efficient count-based exploration in tabular cases.

Note that since $\lambda$ only relates to the overall objective $I_{\rm SD3}$ and does not affect the estimation of state density, the exploration bonus holds for arbitrary $\lambda\geq 1$.

\begin{theorem}\label{thm:2}
In tabular MDPs, optimizing the intra-skill exploration reward is equivalent to count-based exploration, as
\begin{equation}
r^{\rm exp}_{z}(s) \approx \frac{|\cS|/2}{N(s, z)+\kappa}.
\end{equation}
where $N(s, z)$ is the count of visitation of state-skill pair $(s, z)$ in experiences, $|\cS|$ is the total number of states in a tabular case, and $\kappa > 0$ is a small non-negative constant.
\end{theorem}
As a result, maximizing the intra-skill exploration reward is equivalent to performing count-based exploration in previous works \cite{kolter2009near,strehl2008analysis}, which is provable efficient in tabular MDPs \cite{bellemare2016unifying,ostrovski2017count}. Through the approximation in a compact latent space, the intra-skill exploration encourages skill-conditional policy to increase the pseudo-count of rarely visited state-skill pairs in a high-dimensional space. The proof of Theorem~\ref{thm:2} is attached in Appendix~\ref{sec:proof_thm2}.

\begin{figure*}[htbp]
\begin{center}
\centerline{
\includegraphics[width=\textwidth]{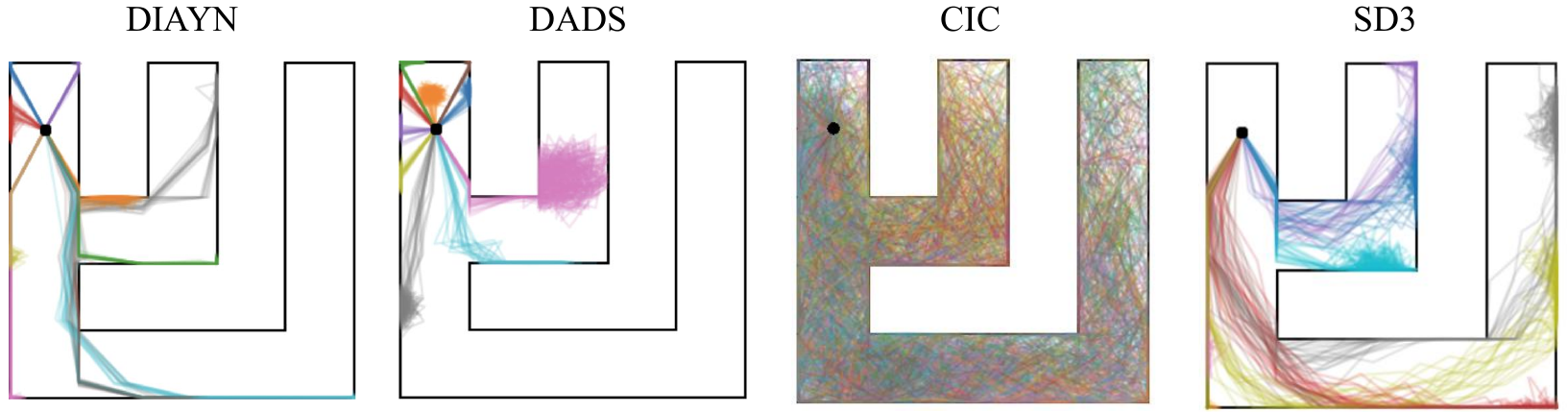}}
\caption{Results for maze experiment. We visually demonstrate the agent's ability to explore the environment and the diversity of skills discovered by the agent. The agent starts from the black dot of the maze and interacts for 250K steps. Both DIAYN and DADS do not reach the right side of the maze while obtaining distinguishable trajectories highlighted by different colors. The trajectories of CIC span the entire maze but appear chaotic. In contrast, SD3 can reach the farthest position from the starting point and facilitates easy differentiation of trajectories of different skills.}
\label{fig:maze_experiment}
\end{center}
\vspace{-2em}
\end{figure*}

\section{Related work}

\subsection{Unsupervised Skill Discovery}
Unsupervised skill discovery in RL aims to acquire a repertoire of useful skills without relying on extrinsic rewards. Early efforts, such as VIC \cite{vic}, DIAYN \cite{diayn}, and DADS \cite{dads}, maximize the MI between the skill and the state to discover diverse skills. However, as noted in EDL \cite{EDL-2020},
LSD \cite{LSD-2022}, and CSD \cite{USD-2023}, such MI-based methods usually prefer static skills caused by poor state coverage and may hinder the application for downstream tasks. Recent methods strive to address this limitation to learn dynamic and meaningful skills. These methods perform explicit exploration or enforce Lipschitz constraints in the representation to maximize the traveled distances of skills. 
% EDL \cite{EDL-2020} separates the exploration and skill discovery process, enabling skill discovery to take place on the basis of discovered states. LSD \cite{LSD-2022} utilizes a Lipschitz constraint to maximize the traveled distances in the state space. 
Further, CIC \cite{cic} employs contrastive learning between state transitions and skills to encourage agent's diverse behaviors. BeCL \cite{becl} uses contrastive learning to differentiate between various behavioral patterns and maximize the entropy implicitly. ReST \cite{recurrent-2022} encourages the trained skill to stay away from the estimated state visitation distributions of other skills. Some methods, like DISCO-DANCE \cite{guidance-2023}, APS \cite{aps}, SMM \cite{SMM} and DISDAIN \cite{disdain}, focus on introducing an auxiliary exploration reward to address insufficient exploration. Furthermore, to verify the effectiveness of skill discovery in large-scale state space (e.g., images), recent methods including Choreographer \cite{Mazzaglia2023Choreographer} and Metra \cite{park2024metra} evaluate the effectiveness of methods on pixel-based URLB \cite{pixel-urlb}, which often relies on model-based agents to learn meaningful knowledge from imagination, and skills are discovered in the latent space. Metra \cite{park2024metra} constructs a latent space associated with the original state space via a temporal distance metric, which enables skill learning in high-dimensional environments by maximizing the coverage. In contrast, our method promotes skill diversity by encouraging deviations in skill density and enhances state coverage through latent space exploration. We validate our approach's efficacy through experiments on state-based and pixel-based tasks across various environments.

\subsection{Unsupervised RL}
According to URLB \cite{URLB}, URL algorithms are classified into three main categories: knowledge-based, data-based, and competence-based. Knowledge-based algorithms \cite{pathak2017curiosity,pathak2019disagreement,burda2018rnd} leverage the agent's predictive capacity or understanding of the environment, and the intrinsic reward is tied to the novelty of the agent's behaviors, encouraging the agent to explore areas where its model is less certain. Data-based algorithms \cite{apt,proto} maximize the state entropy to maximize state coverage of skills. Competence-based algorithms \cite{SMM,diayn,aps,minecraft} pre-train the agent to learn useful skills that can be utilized to complete downstream tasks. Our method can be categorized as competence-based, while also combining the benefit of knowledge-based algorithms to encourage exploration. In addition, some recent algorithms do not easily fit into these categories. For example, LCSD \cite{ju2024rethinking} establishes connections between skills, states, and linguistic instructions to guide task completion based on external language directives. DuSkill \cite{kim2024robust} utilizes a guided diffusion model to generate versatile skills beyond dataset limitations, thereby enhancing the robustness of policy learning across diverse domains. EUCLID \cite{yuan2023euclid} improves downstream policy learning performance by jointly pre-training dynamic models and unsupervised exploration strategies. VGCRL \cite{VGCRL} applies variational empowerment to learn effective state representations, thereby improving exploration.

\begin{figure*}[htbp]
\begin{center}
\centerline{
\includegraphics[width=\textwidth]{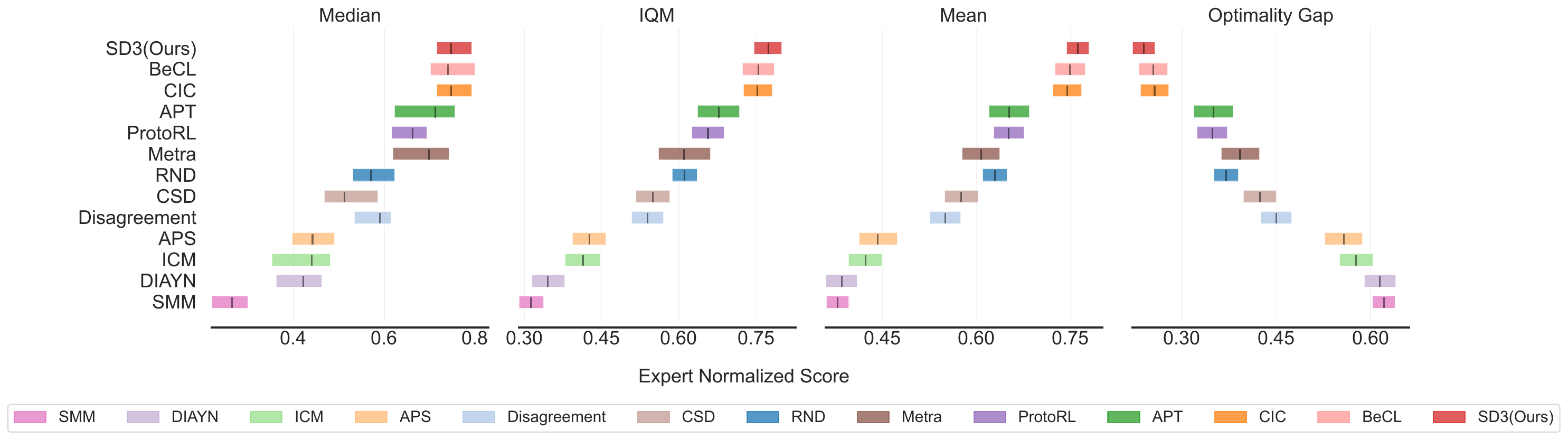}}
\caption{Results for state-based URLB. The aggregate statistics \cite{agarwal2021IQM} indicate the adaptation performance of different unsupervised RL methods in 12 downstream tasks. In terms of IQM, Mean, and OG metrics, SD3 outperforms other competence-based methods and significantly surpasses pure exploration methods, achieving 77.37\%, 76.19\%, and 23.91\%, respectively.}
\label{fig:state_based_experiment}
\end{center}
% \vspace{-1em}
\vspace{-20pt}
\end{figure*}

% Please add the following required packages to your document preamble:
% \usepackage[normalem]{ulem}
% \useunder{\uline}{\ul}{}
\begin{table*}[htbp]
\centering
\caption{Results of SD3 and other baselines on state-based URLB.}
\label{table:numerical_result_state_sd3}
\begin{tabular}{c|cccc|cccc|cccc}
\hline
\multirow{2}{*}{Task} & \multicolumn{4}{c|}{Walker}                                                          & \multicolumn{4}{c|}{Quadruped}                                                        & \multicolumn{4}{c}{Jaco}                                                           \\
                      & Flip                & Run                 & Stand              & Walk                & Jump                & Run                 & Stand               & Walk                & B\_L               & B\_R               & T\_L                & T\_R               \\ \hline
DDPG                  & 538±27              & 325±25              & 899±23             & 748±47              & 236±48              & 157±31              & 392±73              & 229±57              & 72±22              & 117±18             & 116±22              & 94±18              \\ \hline
ICM                   & 390$\pm$10          & 267$\pm$23          & 836$\pm$34         & 696$\pm$46          & 205$\pm$47          & 125$\pm$32          & 260$\pm$45          & 153$\pm$42          & 88$\pm$14          & 99$\pm$8           & 80$\pm$13           & 106$\pm$14         \\
Disagreement          & 332$\pm$7           & 243$\pm$14          & 760$\pm$24         & 606$\pm$51          & 510$\pm$28          & 357$\pm$24          & 579$\pm$64          & 386$\pm$51          & 117$\pm$9          & 122$\pm$5          & 121$\pm$14          & 128$\pm$11         \\
RND                   & 506$\pm$29          & 403$\pm$16          & 901$\pm$19         & 783$\pm$35          & 626$\pm$23          & 439$\pm$7           & 839$\pm$25          & 517$\pm$41          & 102$\pm$9          & 110$\pm$7          & 88$\pm$13           & 99$\pm$5           \\ \hline
APT                   & 606$\pm$30          & 384$\pm$31          & 921$\pm$15         & 784$\pm$52          & 416$\pm$54          & 303$\pm$30          & 582$\pm$67          & 582$\pm$67          & 143$\pm$12         & 138$\pm$15         & 137$\pm$20          & \textbf{170$\pm$7} \\
ProtoRL               & 549$\pm$21          & 370$\pm$22          & 896$\pm$20         & 836$\pm$25          & 573$\pm$40          & 324$\pm$26          & 625$\pm$76          & 494$\pm$64          & 118$\pm$7          & 138$\pm$8          & 134$\pm$7           & 140$\pm$9          \\ \hline
SMM                   & 500$\pm$28          & 395$\pm$18          & 886$\pm$18         & 792$\pm$42          & 167$\pm$30          & 142$\pm$28          & 266$\pm$48          & 154$\pm$36          & 45$\pm$7           & 60$\pm$4           & 39$\pm$5            & 32$\pm$4           \\
DIAYN                 & 361$\pm$10          & 184$\pm$23          & 789$\pm$48         & 450$\pm$37          & 498$\pm$45          & 347$\pm$47          & 718$\pm$81          & 506$\pm$66          & 20$\pm$5           & 17$\pm$5           & 12$\pm$5            & 21$\pm$3           \\
APS                   & 448$\pm$36          & 176$\pm$18          & 702$\pm$67         & 547$\pm$38          & 389$\pm$72          & 201$\pm$40          & 435$\pm$68          & 385$\pm$76          & 84$\pm$5           & 94$\pm$8           & 74$\pm$10           & 83$\pm$11          \\ \hline
CSD                   & {\ul 615$\pm$17}    & 445$\pm$13          & \textbf{962$\pm$7} & 857$\pm$51          & 357$\pm$39          & 362$\pm$60          & 455$\pm$36          & 224$\pm$18          & 99$\pm$7           & 106$\pm$6          & 101$\pm$7           & 154$\pm$11         \\
Metra                 & 600$\pm$48          & 302$\pm$23          & 951$\pm$7          & 756$\pm$67          & 300$\pm$9           & 276$\pm$20          & 637$\pm$85          & 200$\pm$27          & 143$\pm$9          & 142$\pm$8          & 130$\pm$13          & 158$\pm$16         \\
CIC                   & \textbf{641$\pm$26} & {\ul 450$\pm$19}    & {\ul 959$\pm$2}    & {\ul 903$\pm$21}    & 565$\pm$44          & 445$\pm$36          & 700$\pm$55          & 621$\pm$69          & \textbf{154$\pm$6} & {\ul 149$\pm$4}    & \textbf{149$\pm$10} & {\ul 163$\pm$9}    \\
BeCL                  & 611$\pm$18          & 387$\pm$22          & 952$\pm$2          & 883$\pm$34          & \textbf{727$\pm$15} & \textbf{535$\pm$13} & \textbf{875$\pm$33} & {\ul 743$\pm$68}    & 148$\pm$13         & 139$\pm$14         & 125$\pm$10          & 126$\pm$10         \\ \hline
SD3(Ours)                   & 595$\pm$25          & \textbf{451$\pm$23} & 930$\pm$5          & \textbf{914$\pm$11} & {\ul 676$\pm$29}    & {\ul 471$\pm$13}    & {\ul 847$\pm$17}    & \textbf{752$\pm$40} & {\ul 151$\pm$7}    & \textbf{152$\pm$9} & {\ul 142$\pm$7}     & 152$\pm$7          \\ \hline        
\end{tabular}
\end{table*}

\section{Experiments}
\label{sec:experiments}

We start by introducing experiments in Maze to visualize the skills. Subsequently, we validate the effectiveness of SD3 by conducting experiments on challenging tasks from the DeepMind Control Suite (DMC) \cite{dmc}, with both state-based \cite{URLB} and pixel-based \cite{pixel-urlb} observations. Finally, we conduct ablation studies to demonstrate the factors that influence the effectiveness of SD3. 

\subsection{Maze Experiment}

We conduct experiments in a 2D maze to visually demonstrate the learned skills, as shown in Fig.~\ref{fig:maze_experiment}. The agent's initial state is represented by a black dot, with different colored lines indicating the trajectories corresponding to the different skills it has learned. The agent's state is the current positional information, and the actions represent the velocity and direction of movement. Building on this, we compare SD3 with two classical MI-based methods, DIAYN \cite{diayn} and DADS \cite{dads}, whose objectives correspond to the reverse form $\mathcal{H}(Z)-\mathcal{H}(Z|S)$ and the forward form $\mathcal{H}(S)-\mathcal{H}(S|Z)$ of the MI term $I(S;Z)$, respectively. Additionally, we compare SD3 with an entropy-based CIC algorithm \cite{cic}, whose primary objective is to maximize state-transition entropy $\mathcal{H}(\tau)$ to generate diverse behaviors. We employ the PPO as the backbone and train $n = 10$ skills for each algorithm.

We delineate the learned skills of each algorithm within the maze environment in Fig.~\ref{fig:maze_experiment} and introduce two key metrics for comparing SD3 with other methods: state coverage and distinguishability of skills, where insufficient state coverage may impede the acquisition of dynamic skills, and the lack of distinguishability leads to similar behaviors of skills. 
% which pertains to the extent of divergence among trajectories characterized by distinct colors for each algorithm is equally important, as it relates to the diversity of skills. 
% As illustrated in Figure \ref{fig:maze_experiment}, 
According to the results,
(\romannumeral1) DIAYN and DADS fail to extend to the upper-right corner of the maze, but exhibit clear distinctions among trajectories of skills, indicating that merely maximizing $I(S;Z)$ can learn discriminable skills but lack effective exploration of the state space; (\romannumeral2) CIC demonstrates the best state coverage while learns skills with mixed trajectories due to the maximization of $\cH(s)$ as its primary objective; (\romannumeral3) In contrast, SD3 strikes a balance between state coverage and empowerment in skill discovery. It learns discriminable skills by maximizing the deviation between the state densities of a certain skill and others. Meanwhile, SD3 achieves commendable state coverage through latent space exploration. 

% \begin{figure*}[!t]
% \centering
% \subfloat[]{\includegraphics[width=3in]{Fig/exp_pixel.pdf}%
% \label{fig:exp_pixel}}
% \hfil
% \subfloat[]{\includegraphics[width=3in]{Fig/robustness.pdf}%
% \label{fig:robustness}}
% \caption{(a) Results for Pixel-based URLB. We conduct experiments on pixel-based URLB to demonstrate the scalability of SD3 for large-scale problems. (b) Results for robustness experiment. It can be observed that SD3 retains higher performance ratio than CIC in the noisy domain.}
% \label{fig:exp}
% \vspace{-1em}
% \end{figure*}

\begin{figure}[!t]
\centering
\includegraphics[width=0.5\textwidth]{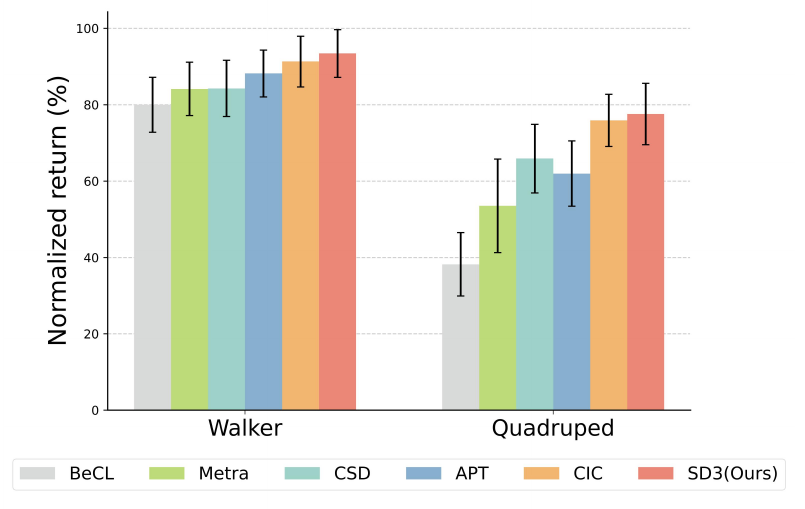}
\caption{Results for Pixel-based URLB. We conduct experiments on pixel-based URLB to demonstrate the scalability of SD3 for large-scale problems.}
\label{fig:exp_pixel}
\vspace{-1em}
\end{figure}

\begin{figure}[!t]
\centering
\includegraphics[width=0.45\textwidth]{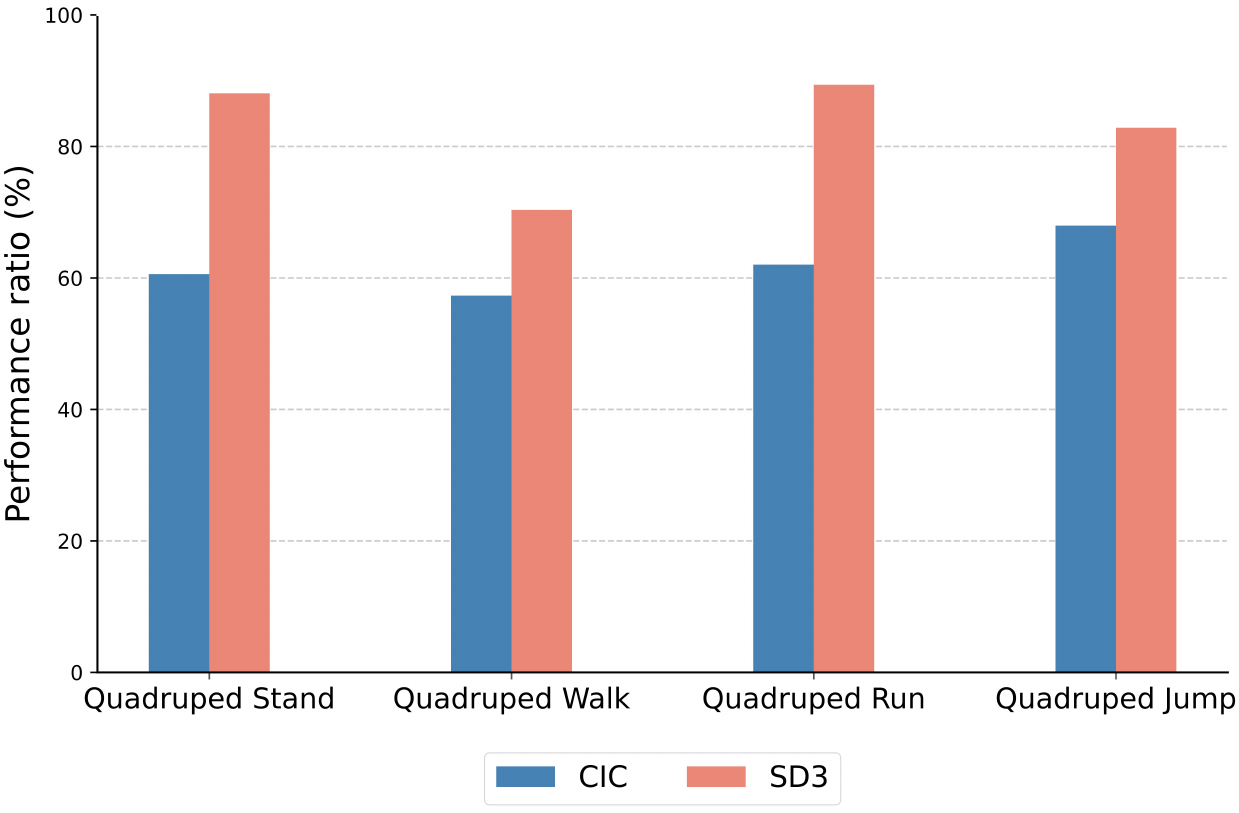}
\caption{Results for robustness experiment. It can be observed that SD3 retains higher performance ratio than CIC in the noisy domain.}
\label{fig:exp}
\vspace{-1em}
\end{figure}

\subsection{State-based URLB}
\label{sec:state-based}
According to state-based URLB \cite{URLB}, we evaluate our approaches in 12 downstream tasks across 3 distinct continuous control domains, each designed to evaluate the effectiveness of algorithms under high-dimensional state spaces. The three domains are \emph{Walker}, \emph{Quadruped}, and \emph{Jaco Arm}. Specifically, \emph{Walker} involves a biped constrained to a 2D vertical plane with a state space $\cS \in \RR^{24}$ and an action space $\cA \in \RR^{6}$. The agent in the \emph{Walker} domain must learn to maintain balance and move forward, completing four downstream tasks: \emph{stand, walk, run,} and \emph{flip}. \emph{Quadruped} features a four-legged robot in a 3D environment, characterized by a state space $\cS \in \RR^{78}$ and an action space $\cA \in \RR^{16}$. The downstream tasks, including \emph{stand, run, jump,} and \emph{walk}, pose challenges to the agent due to the complex dynamics of its movements. \emph{Jaco} employs a 6-DOF robotic arm with a three-finger gripper, functioning within a state space $\cS \in \RR^{55}$ and an action state $\cA \in \RR^{9}$. Primary downstream tasks in \emph{Jaco} Arm include reaching and manipulating objects at various positions.
% and assessing the agent’s precision and adaptability in complex manipulations. 

\textbf{Baselines.} We conduct comparisons between SD3 and the baselines delineated across the three URL algorithm categories as defined by URLB \cite{URLB}. These categories encompass knowledge-based baselines, which consist of ICM \cite{pathak2017curiosity}, Disagreement \cite{pathak2019disagreement}, and RND \cite{burda2018rnd}; data-based baselines, which include APT \cite{apt} and ProtoRL \cite{proto}; and competence-based baselines, comprising SMM \cite{SMM}, DIAYN \cite{diayn}, and APS \cite{aps}. Furthermore, we extend our comparisons to include other novel competence-based algorithms such as CSD \cite{USD-2023}, Metra \cite{park2024metra}, BeCL \cite{becl}, and CIC \cite{cic}.

\textbf{Evaluation.}
We employ a rigorous evaluation to assess the performance of SD3 alongside other algorithms, involving a two-phase process. Initially, a pre-training of 2M steps is performed using only intrinsic rewards, followed by a fine-tuning phase of 100K steps on each downstream task using extrinsic rewards. Building upon prior work \cite{URLB}, we utilize DDPG as the backbone algorithm.
To ensure statistical rigor and mitigate the impact of incidental factors in RL training, we conduct experiments across multiple seeds ($10$ seeds per algorithm), resulting in a substantial volume of runs (i.e., $1560$ $=$ $13$ algorithms $\times$ $10$ seeds $\times$ $3$ domains $\times$ $4$ tasks). The detailed scores in downstream tasks are attached in Table~\ref{table:numerical_result_state_sd3}. We employ four statistical metrics to assess performance: Median, interquatile mean (IQM), Mean, and optimality gap (OG) \cite{agarwal2021IQM}. IQM focuses on the central tendency of the middle 50\%, excluding the top and bottom quartiles. OG understands the extent to which the algorithm approaches the optimal level, where the optimal level is determined by the expert models' ultimate score obtained on each downstream task. 
% Mean is utilized to evaluate the average performance of each algorithm across various downstream tasks. Median provides a direct measure of central tendency less susceptible to extreme value influence. 
% IQM focuses on the central tendency of the middle 50\% of the data, excluding the top and bottom quartiles, aiding in further reducing the impact of outliers. OG helps us understand the extent to which algorithm performance approaches the optimal level, where the optimal level is determined by the expert models' ultimate score obtained on each downstream task.

\textbf{Results.}
According to Fig.~\ref{fig:state_based_experiment},
% We standardize each score against the expert score, and the statistical findings are depicted in . 
% on the , 
SD3 achieves the highest IQM score at 77.37\%, slightly surpassing CIC and BeCL, which scores 75.19\% and 75.38\% respectively, and significantly outperforming other competence-based algorithms such as Metra (61.01\%), CSD (54.93\%), and APS (43.61\%). On the OG metric, SD3's gap to optimal performance is 23.91\%, marginally better than CIC and BeCL at 25.65\% and 25.44\%, respectively, and far superior to Metra (39.25\%), CSD (42.43\%), and APS (55.76\%). Additionally, compared to purely exploratory methods, SD3 significantly outperforms the best-performing method, APT, on both IQM and OG metrics, with APT scoring 67.74\% and 34.98\% on these metrics, respectively.
% In our analysis, 
The remarkable performance of SD3 stems from two main factors. First, the use of $r^{\rm sd3}$ facilitates the learning of distinguishable skills by the agent, thereby facilitating effective adaptation across various downstream tasks. Second, the learned compressed representation of the high-dimensional state space leads to efficient intra-skill exploration within a compact space, which not only maintains skill consistency but also enhances exploration ability. 

\begin{table*}[htbp]
\centering
\caption{Results of SD3 and baselines on pixel-based URLB.}
\label{table:numerical_result_pixel_sd3}
\resizebox{0.9\textwidth}{!}{
\tiny
\begin{tabular}{cc|ccccc|c}
\hline
Domain                     & Task  & APT             & CSD        & Metra      & CIC                & BeCL       & SD3(Ours)                 \\ \hline
\multirow{4}{*}{Walker}    & Flip  & 803$\pm$26      & 681$\pm$56 & 665$\pm$32 & {\ul 836$\pm$12}   & 539$\pm$8  & \textbf{864$\pm$27} \\
                           & Run   & {\ul 506$\pm$4} & 451$\pm$41 & 454$\pm$29 & 504$\pm$21         & 456$\pm$14 & \textbf{543$\pm$22} \\
                           & Stand & 961$\pm$5       & 958$\pm$13 & 968$\pm$4  & {\ul 973$\pm$2}    & 968$\pm$4  & \textbf{982$\pm$1}  \\
                           & Walk  & 880$\pm$37      & 948$\pm$5  & 949$\pm$3  & \textbf{953$\pm$5} & 939$\pm$1  & {\ul 945$\pm$3}     \\ \hline
\multirow{4}{*}{Quadruped} & Jump  & 557$\pm$67      & 580$\pm$74 & 677$\pm$27 & {\ul 723$\pm$16}   & 340$\pm$32 & \textbf{729$\pm$16} \\
                           & Run   & 396$\pm$9       & 390$\pm$21 & 276$\pm$46 & \textbf{439$\pm$3} & 162$\pm$4  & {\ul 438$\pm$16}    \\
                           & Stand & 785$\pm$18      & 854$\pm$20 & 788$\pm$29 & {\ul 873$\pm$13}   & 583$\pm$56 & \textbf{921$\pm$3}  \\
                           & Walk  & 475$\pm$55      & 530$\pm$19 & 181$\pm$39 & {\ul 672$\pm$15}   & 283$\pm$39 & \textbf{680$\pm$43} \\ \hline
\end{tabular}}
\end{table*}

\begin{table*}[htbp]
\centering
\caption{Results of robustness experiments.}
\label{table:robust_exp}
\resizebox{0.9\textwidth}{!}{
\tiny
\begin{tabular}{lc|ccc|ccc}
\hline
Domain                     & Task  & \begin{tabular}[c]{@{}c@{}}CIC\\ (Noisy)\end{tabular} & \begin{tabular}[c]{@{}c@{}}CIC\\ (Normal)\end{tabular} & \begin{tabular}[c]{@{}c@{}}Performance \\ Ratio\end{tabular} & \begin{tabular}[c]{@{}c@{}}SD3\\ (Noisy)\end{tabular} & \begin{tabular}[c]{@{}c@{}}SD3\\ (Normal)\end{tabular} & \begin{tabular}[c]{@{}c@{}}Performance\\ Ratio\end{tabular} \\ \hline
\multirow{4}{*}{Walker}    & Flip  & 511$\pm$6                                             & 641$\pm$26                                             & 79.72\%                                                      & 554$\pm$24                                            & 595$\pm$25                                             & \textbf{93.11\%}                                            \\
                           & Run   & 319$\pm$20                                            & 450$\pm$19                                             & 70.89\%                                                      & 330$\pm$25                                            & 451$\pm$23                                             & \textbf{73.17\%}                                            \\
                           & Stand & 845$\pm$12                                            & 959$\pm$2                                              & 88.11\%                                                      & 909$\pm$11                                            & 930$\pm$5                                              & \textbf{97.74\%}                                            \\
                           & Walk  & 784$\pm$46                                            & 903$\pm$21                                             & 86.82\%                                                      & 877$\pm$27                                            & 914$\pm$11                                             & \textbf{95.95\%}                                            \\ \hline
\multirow{4}{*}{Quadruped} & Jump  & 384$\pm$61                                            & 565$\pm$44                                             & 67.96\%                                                      & 560$\pm$48                                            & 676$\pm$29                                             & \textbf{82.84\%}                                            \\
                           & Run   & 276$\pm$48                                            & 445$\pm$36                                             & 62.02\%                                                      & 421$\pm$47                                            & 471$\pm$13                                             & \textbf{89.38\%}                                            \\
                           & Stand & 424$\pm$25                                            & 700$\pm$55                                             & 60.57\%                                                      & 746$\pm$93                                            & 847$\pm$17                                             & \textbf{88.07\%}                                            \\
                           & Walk  & 356$\pm$99                                            & 621$\pm$69                                             & 57.32\%                                                      & 529$\pm$55                                            & 752$\pm$40                                             & \textbf{70.34\%}                                            \\ \hline
\multicolumn{2}{c|}{Average}       & --                                                    & --                                                     & 71.68\%                                                      & --                                                    & --                                                     & \textbf{86.33\%}                                            \\ \hline
\end{tabular}}
\end{table*}

\subsection{Pixel-based URLB}
To further validate the effectiveness of SD3, we conduct experiments on pixel-based URLB \cite{pixel-urlb}, which includes \emph{Walker} and \emph{Quadruped} domains with 8 downstream tasks. The pixel-based environment employs raw pixel data as input, foregoing abstracted features, or processed sensor information. The challenge of deriving meaningful skills from such unrefined inputs is substantial, particularly in the absence of external rewards. Meanwhile, exploration becomes more difficult in image-based spaces, thereby testing the exploration ability of algorithms under conditions that closely resemble practical applications. 

\textbf{Baselines.}
We compared SD3 with the top three performing algorithms in state-based experiments, i.e., BeCL \cite{becl}, CIC \cite{cic}, and APT \cite{apt}, as well as with the recently proposed skill discovery algorithms including CSD \cite{USD-2023} and Metra \cite{park2024metra}. Among these, APT stands out as a data-based algorithm, which can also be considered a representative of pure exploration algorithms and demonstrates strong performance in exploring environments. The others are competence-based algorithms, which accomplish downstream tasks by learning useful and diverse skills. 
% The primary distinction among them lies in their intrinsic reward objectives.

\textbf{Evaluation.} 
We conduct 2M steps of pre-training solely based on intrinsic rewards in each domain, followed by 100K steps of fine-tuning on the downstream tasks using extrinsic rewards. The scores achieved in the downstream tasks are used to evaluate the algorithm. According to the official benchmark of the pixel-based URLB \cite{pixel-urlb},  unsupervised RL algorithms often perform poorly when combined with a model-free method (e.g., DDPG \cite{ddpg} or DrQv2 \cite{DrQv2}) with image observations, while performing much better when using a model-based backbone (e.g., Dreamer \cite{Hafner2021DreamerV2}). Thus, we follow this setting and conduct experiments with Dreamer backbone. We report the average adaptation performance in Fig.~\ref{fig:exp}(a). In the relatively simple \emph{Walker} domain, SD3 achieves the best performance (93.42\%), slightly outperforming other methods (i.e., CIC-91.29\%, APT-88.17\%, CSD-84.26\%). 
% which suggests that while SD3 excels in the Walker domain, simpler domain may not fully differentiate the performance of various algorithms. 
% In contrast, 
In the challenging Quadruped domain, SD3 outperforms CIC (77.57\% and 75.89\%, respectively) and shows significant improvement over other competence-based methods (i.e., CSD-65.89\%, Metra-53.53\%) and the best pure-exploration method in state-based URLB (i.e., APT-61.96\%). This highlights SD3's commendable advantages in both various image-based tasks. The detailed scores are attached in Table~\ref{table:robust_exp}.

\subsection{Robustness Experiment}
\label{sec:robust_exp}

Unlike CIC, APS, and BeCL, which rely on entropy-based exploration strategies, SD3 introduces a novel exploration reward that resembles a UCB-style bonus. Such a UCB-term in exploration is provable efficient in linear and tabular MDPs, which has been rigorously studied in previous research~\cite{lsvi-ucb,zhang2021rewardfree}. 
%Under the reward-free setting of this work, the UCB-based exploration is provable to collect data with sufficiently good coverage of the underlying MDP~\cite{zhang2021rewardfree}, which is essential for direct policy learning or fast adaptation with specific rewards. 
In contrast, the entropy-based exploration used in previous methods has the disadvantage of being non-robust (e.g., adding small noise will significantly affect its entropy). Thus, to further verify that the robustness of SD3, we conduct experiments in noisy domains of URLB by adding noise during pre-training, which is sampled from $N(0,0.1)$, followed by noise-free fine-tuning to assess the learned skills. 

\textbf{Evaluation.} We choose CIC for comparison, which performs competitively with our method in standard URLB. Each technique is evaluated across 5 random seeds and the results are given in Fig.~\ref{fig:exp}(b). The Performance Ratio (PR) denotes the ratio of the adaptation score in the noisy domain to that in the normal setup. According to the results, it is evident that the UCB-bonus used in SD3 is more robust than entropy-based rewards in noisy environments, achieving significantly higher Performance Ratio than CIC. The detailed results are attached in Table~\ref{table:robust_exp}.

\subsection{Ablation Studies}
We provide ablation studies for components in skill discovery and skill adaptation of SD3. For skill discovery, we perform the comparison on (\romannumeral1) density estimation with and without soft modularization.
% and (\romannumeral2) the different settings of temperatures in the routing network. 
The final rewards for skill discovery contain $r^{\rm sd3}_z(s)$ and $r^{\rm exp}_z(s)$. We conduct ablation studies on (\romannumeral2) different settings of $\lambda$ in calculating $r^{\rm sd3}_z(s)$, as well as (\romannumeral3) the different balance factors of the two rewards. For skill adaptation, we sampled skills randomly to evaluate their generalization ability in our main results. In ablation studies, (\romannumeral4) we evaluate two more skill-choosing strategies in adaptation for a comparison.
% We also provide visualizations of skills learned in tree-like Maze and DMC tasks. The results show that SD3 learns dynamic and valuable skills, enabling the agent to adapt to downstream tasks quickly. 

\begin{figure}[!t]
\centering
\includegraphics[width=0.45\textwidth]{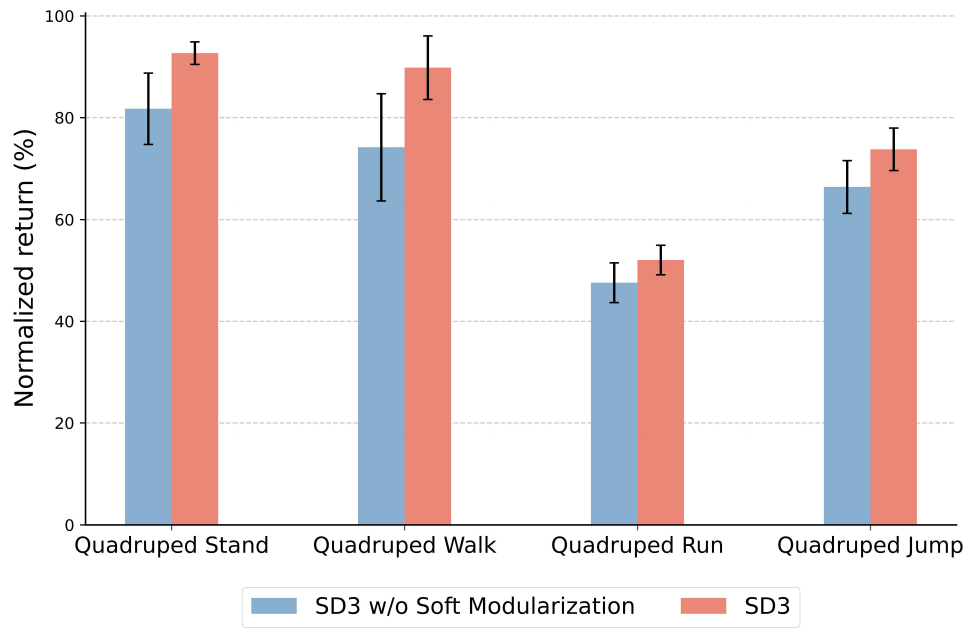}
\caption{Ablation on the soft modularization structure.}
\label{fig:abl_soft_modul}
\vspace{-1em}
\end{figure}

\subsubsection{Impact of Soft Modularization}
\label{sec:ablation_soft_modu}
As mentioned in section \ref{sec:skill-discovery}, we use CVAE to estimate the state density of different skills. To enhance the accuracy of estimation in complex state spaces, we have introduced soft modularization into the traditional CVAE structure. Consequently, we conduct an ablation study on the soft modularization.
% to test the performance of SD3 with and without soft modularization. 
Aggregated scores are reported in Fig.~\ref{fig:abl_soft_modul}. We observe that SD3 with soft modularized CVAE obtains superior performance,
% always beneficial, 
as it has sufficient capacity to learn the density information of different skills for the same state in complex state spaces, while the skill density estimation of one skill may intervene with those of other skills in the traditional CVAE.

\begin{figure}[!t]
\centering
\includegraphics[width=0.4\textwidth]{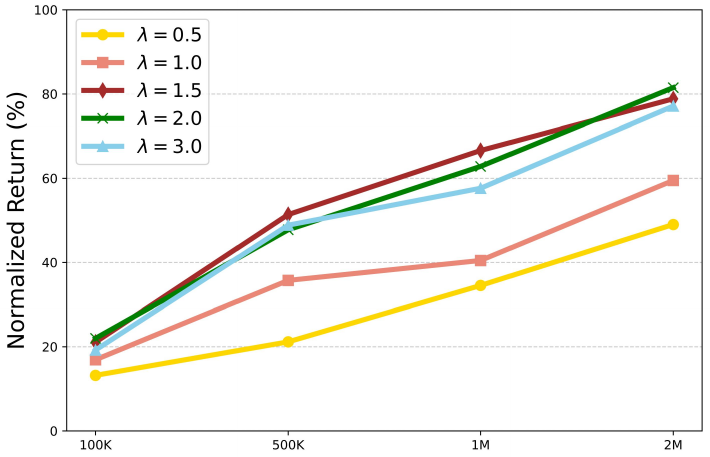}
\caption{Results for the impact of weight parameter in the \emph{Quadruped}. When $\lambda$ is set to 0.5 or 1, SD3 performs poorly. However, it is observed that increasing lambda beyond 1 does not significantly impact the performance of SD3.}
\label{fig:abl_lambda}
\vspace{-1em}
\end{figure}

\subsubsection{Impact of Weight Parameter $\lambda$}
The discussion in section \ref{sec:skill-discovery} introduces a weight parameter $\lambda$ in Eq.(\ref{eq:sd3-1}). %As we have discussed, we introduce a weight parameter $\lambda$ in Eq(\ref{eq:sd3-1}). 
To investigate the impact of $\lambda$, we conduct an ablation study by varying $\lambda$ from $[0.5, 1.0, 1.5, 2.0, 3.0]$. The results, exhibited in Fig.~\ref{fig:abl_lambda}, indicate that the performance of SD3 fluctuates within a narrow range when lambda is greater than 1. Therefore, we conclude that $\lambda$ is generally applicable in a wider range, and SD3 is not sensitive to the parameter when $\lambda>=1.5$.

\begin{figure*}[htbp]
\begin{center}
\centerline{
\includegraphics[width=0.95\textwidth]{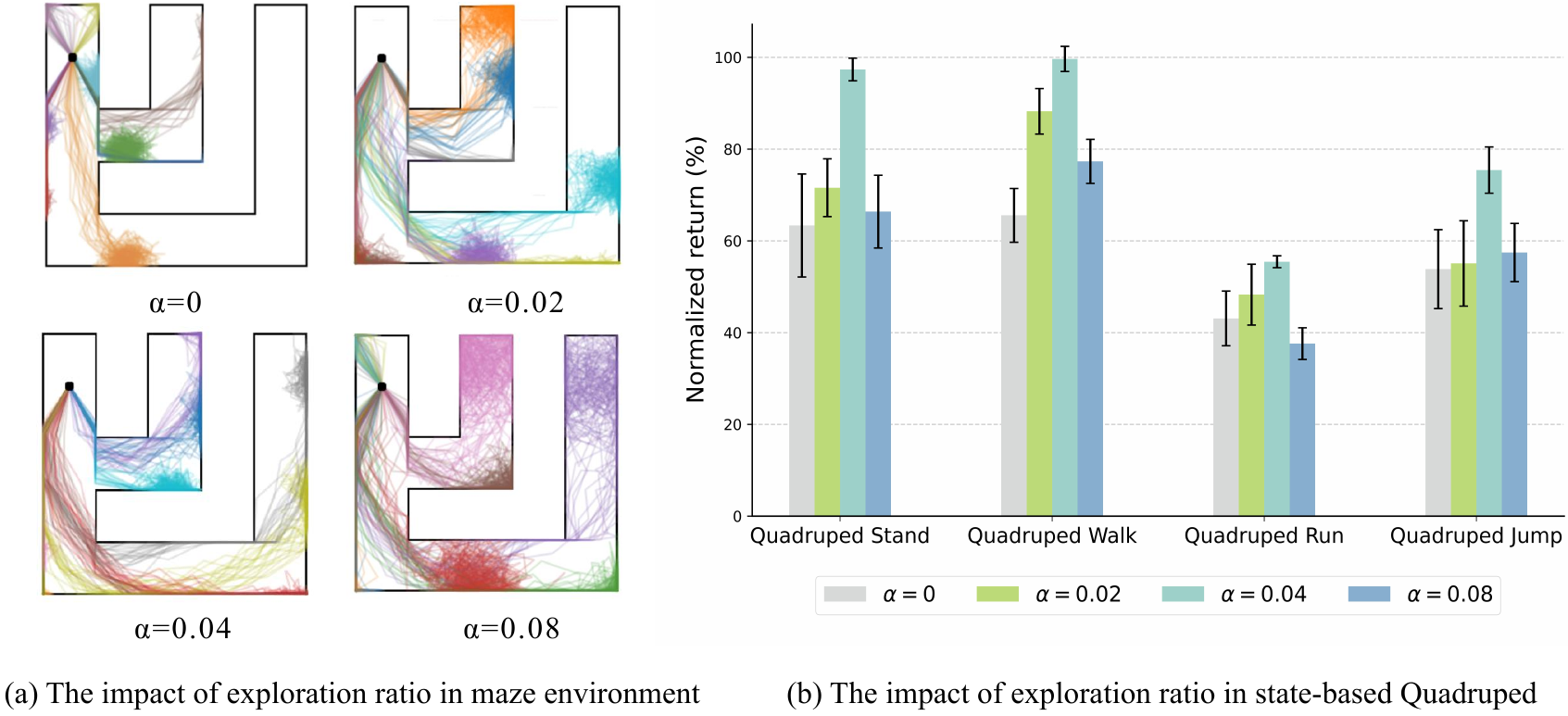}}
\caption{Results for the impact of exploration ratio. (a) We conduct experiments with different $\alpha$ in the maze and found that varying $\alpha$ values significantly impact both the state coverage and the stability of learned skills. (b) In the \emph{Quadruped} domain, different $\alpha$ also have a notable effect on the performance of various downstream tasks.}
\label{fig:ablation_ratio}
\end{center}
\vspace{-2em}
\end{figure*}

\subsubsection{The Exploration Ratio}
We conduct an ablation on the different exploration ratios $\alpha$,
% that affect the performance of SD3. 
Specifically, with the hyper-parameter $\alpha$, the reward is represented as:

\begin{equation}
\label{eq:rew-intr}
r^{\rm total}_{z}(s)=r^{\rm sd3}_{z}(s) + \alpha \cdot r^{\rm exp}_{z}(s).
\end{equation}

As illustrated in Fig.~\ref{fig:ablation_ratio}(a), when $\alpha$ is set to $0$ and $0.02$, the agent can learn distinguishable and convergent skills but fails to fully explore the maze. When $\alpha$ is set to $0.08$, the agent explores sufficiently, but the trajectories at the endpoints are quite scattered, indicating that the learned skill strategies lack stability. In contrast, $\alpha=0.04$ balances exploration and skill diversity. According to our analysis, when the proportion of exploration is deficient or even absent, SD3 solely maximize $I_{\rm SD3}$. Conversely, an excessively high $\alpha$ can overly prioritize intra-skill exploration, resulting in instability within the learned skills. Empirically, $\alpha = 0.04$ can lead to promising results in downstream tasks in the \emph{Quadruped} domain.

\begin{figure}[!t]
\centering
\includegraphics[width=0.45\textwidth]{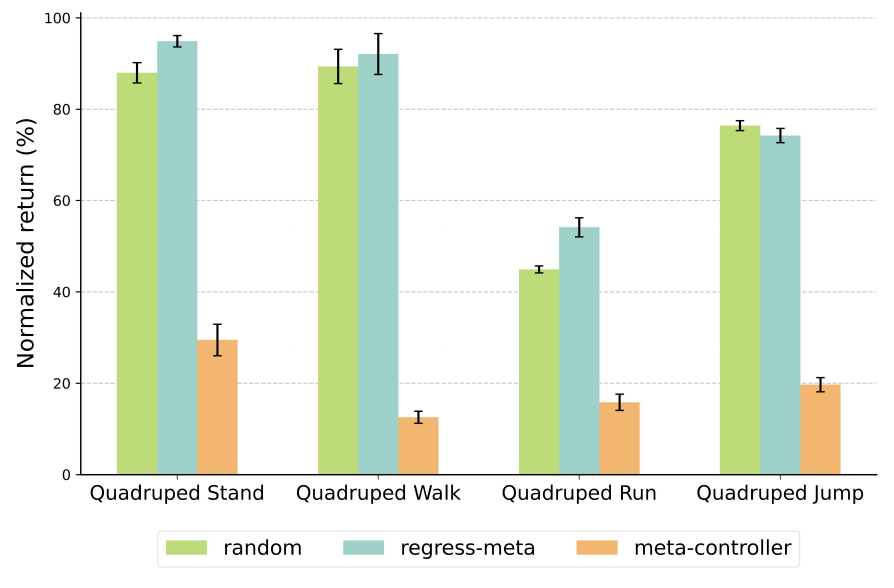}
\caption{Skill adaption strategies ablation. We test several adaptation methods in the fine-tuning phase and find that randomly selecting skills perform comparably to using regress-meta, but employing the meta-controller results in a decline in the performance.}
\label{fig:abl_skill_choose}
\vspace{-1em}
\end{figure}

\subsubsection{Skill Adaption Strategies in Fine-tuning}
% Previous work \cite{URLB} has shown that during the fine-tuning phase, performance across different skills does not always level equally; some skills demonstrate weaker adaptability in downstream tasks, while others show the opposite. Therefore, we investigate various skill adaptation methods in the state-based environment to assess their impact on algorithm performance in downstream tasks. 

In the experiment described in section \ref{sec:state-based}, we follow the URLB standards for fair comparison, employing random skill sampling during fine-tuning to evaluate average skill performance. To enhance skill adaptation, we introduce two methods: regress-meta and meta-controller. Regress-meta estimates the expected reward of each skill during the initial 4K fine-tuning steps to compute its skill-value, selecting the skill with the highest value for downstream tasks. Meta-controller trains a high-level controller, $\mu(z|s)$, during fine-tuning to select the most suitable skill $z$ based on the current state $s$. This controller integrates with the pre-trained policy $\pi(a|s,z)$ to optimize the high-level policy, defined as $\pi(a|s) = \sum_{z \in \mathcal{Z}} \mu(z|s) \pi(a|s,z)$.

The results in Fig.~\ref{fig:abl_skill_choose} demonstrate that regress-meta improves performance over random skill selection in \emph{Quadruped Stand}, \emph{Walk}, and \emph{Run}, but shows a slight decline in \emph{Quadruped Jump}. This outcome likely stems from regress-meta's strategy of consistently selecting the skill with the highest expected reward during the initial fine-tuning phase. While this increases the likelihood of selecting a well-adapted skill, it may also favor skills that perform well in the first 4K steps but underperform in later stages. Conversely, the meta-controller shows comparatively poor performance, which we attribute to its reliance on large amounts of training data, making it challenging to converge within the 100K fine-tuning steps.

\begin{figure*}[htbp]
\begin{center}
\centerline{
\includegraphics[width=0.95\textwidth]{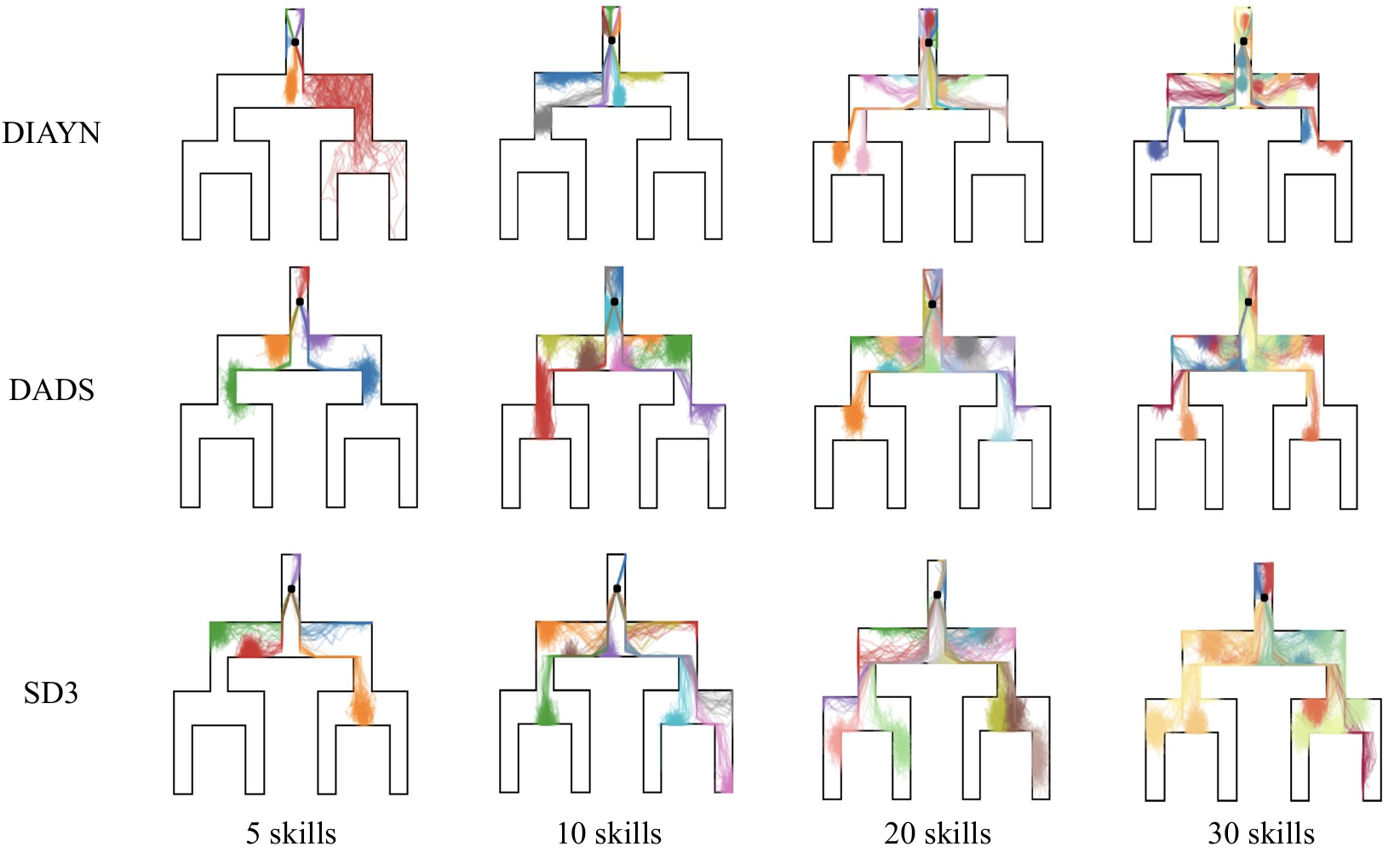}}
\caption{Additional experiments in the tree-like Maze with different numbers of skills. Under different environmental conditions, SD3 demonstrates superior exploration capabilities while still learning distinguishable skills, outperforming DIAYN and DADS.}
\label{fig:tree_mze}
\end{center}
\vspace{-2em}
\end{figure*}

\subsection{Visualization}
\subsubsection{Tree-like Maze}
As shown in Fig.~\ref{fig:tree_mze}, we conduct additional experiments in the tree-like maze to visualize the skills learned by SD3. It can be observed that DIAYN and DADS only reach the middle of the maze, whereas SD3 successfully reaches the bottom of the maze. The proposed latent space reward in SD3 demonstrates strong exploration ability in large-scale mazes. Moreover, the trajectories of different skills remain distinguishable in SD3.

\begin{figure*}[htbp]
\centering
\includegraphics[width=0.9\textwidth]{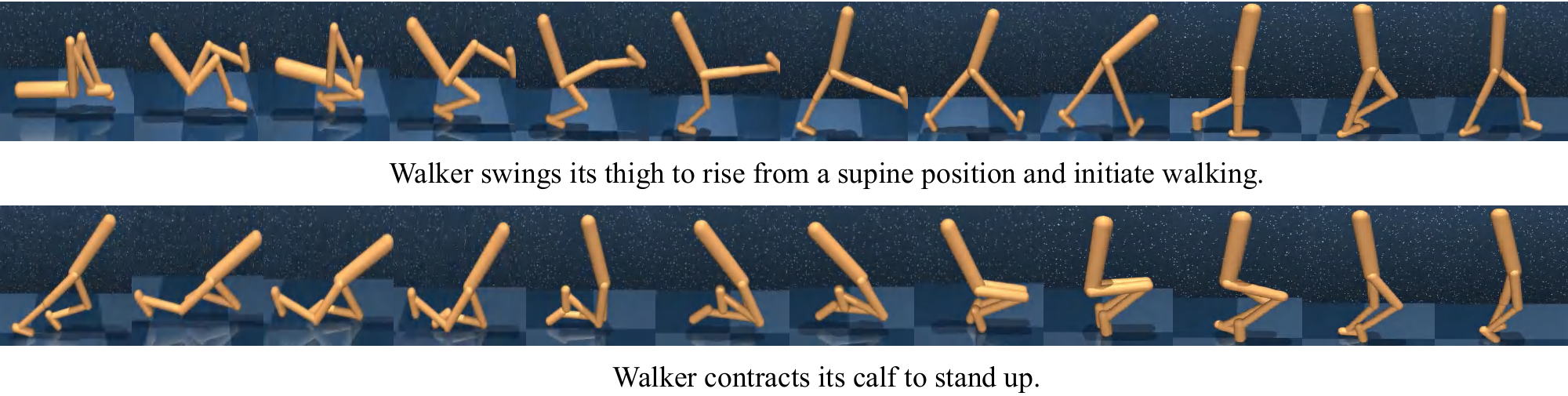}\\
\includegraphics[width=0.9\textwidth]{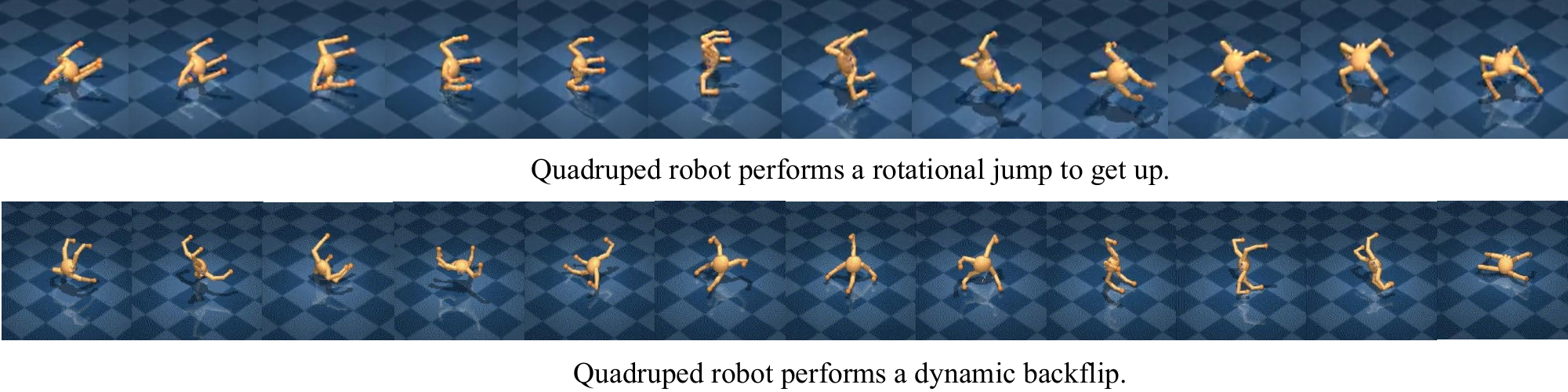}\\
\includegraphics[width=0.9\textwidth]{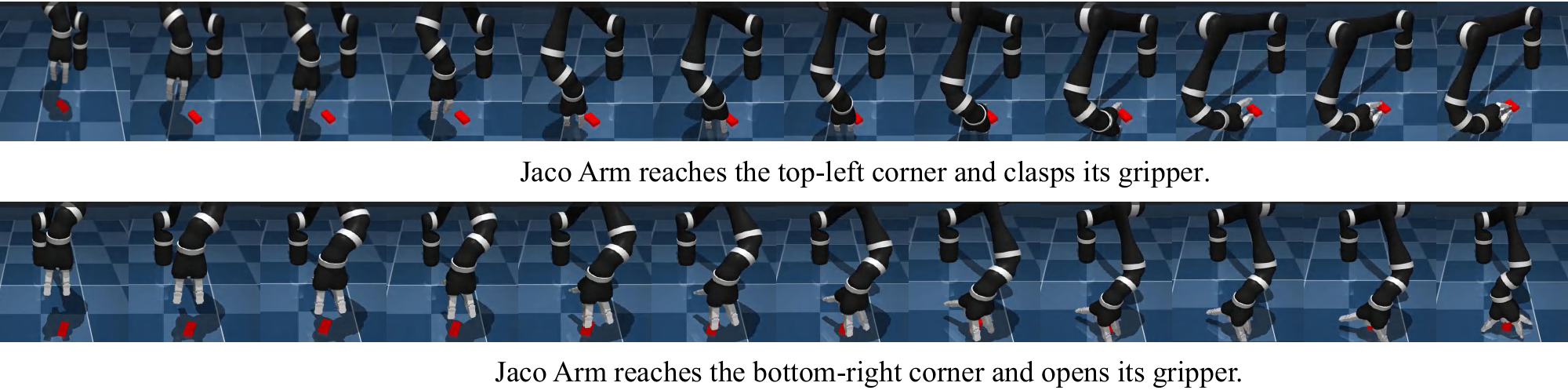}
\caption{Skill visualization in DMC. It can be observed that SD3 learns dynamic and valuable skills, which enable the agent to quickly adapt to downstream tasks.}
\label{fig:dmc_visual}
\end{figure*}

\subsubsection{Deepmind Control Suite}
Fig.~\ref{fig:dmc_visual} shows the learned skills in the \emph{Walker}, \emph{Quadruped}, and \emph{Jaco Arm} domains. The result shows SD3 can learn various locomotion skills, including standing, walking, rolling, moving, and somersault; and also learns various manipulation skills by moving the arm to explore different areas, opening and closing the gripper in different locations. The learned meaningful skills lead to superior generalization performance in the fine-tuning stage of various downstream tasks.

\section{Conclusion}

We propose a novel skill discovery method that promotes skill diversity by encouraging skill deviations in state density and enhancing state coverage through latent space exploration. We realize a novel soft modularization architecture for state density estimation of different skills. Theoretically, the skill discovery objective also maximizes the initial MI term, and the resulting intra-skill exploration bonus resembles count-based exploration. 
% We have showed the effectiveness of SD3 in learning meaningful skills and superior performance in skill adaptation on maze and URLB.
Moreover, our four experiments complement each other and collectively provide sufficient evidence that SD3 demonstrates superior and more comprehensive performance compared to other methods.
% including the ability to discover distinguishable skills (i.e., in maze/URLB domains), superior performance in downstream tasks (i.e., in state/pixel URLB), and scalability to large-scale problems (i.e., pixel-based domains).
One limitation of our method is that the soft modularization architecture is limited to discrete skill spaces, and the theoretical analysis of the exploration bonus requires the assumption of tabular MDPs. In the future, we will extend the idea of skill discovery to LLM-based agents to learn meaningful skills in more complex environments.

% {\appendix[Proof of the Zonklar Equations]
% Use $\backslash${\tt{appendix}} if you have a single appendix:
% Do not use $\backslash${\tt{section}} anymore after $\backslash${\tt{appendix}}, only $\backslash${\tt{section*}}.
% If you have multiple appendixes use $\backslash${\tt{appendices}} then use $\backslash${\tt{section}} to start each appendix.
% You must declare a $\backslash${\tt{section}} before using any $\backslash${\tt{subsection}} or using $\backslash${\tt{label}} ($\backslash${\tt{appendices}} by itself
%  starts a section numbered zero.)}

 % \appendix

% {\appendices
% \section*{Proof of the First Zonklar Equation}
% Appendix one text goes here.
% You can choose not to have a title for an appendix if you want by leaving the argument blank
% \section*{Proof of the Second Zonklar Equation}
% Appendix two text goes here.}

\appendices
\section{Proof of Eq.~\ref{eq:gradient}}
\label{sec:proof_grad}
\begin{proof}
As we discussed in~\ref{sec:skill-discovery}, since we uniformly sample skills from the skill set that contains $n$ skills, we have $p(z)=1/n$ for each skill. Then we have
\begin{equation}
\begin{aligned}
        I_{\mathrm{SD3}}(s,z) &=\log\frac{\lambda n d_{z}^{\pi}(s)}{\lambda d_{z}^{\pi}(s)+\sum_{z^{\prime}\neq z}d_{z^{\prime}}^{\pi}(s)}\\
        &=\log\frac{\lambda n d_{z}^{\pi}(s)}{\lambda d_{z}^{\pi}(s)+\rho_{z^{c}}(s)}.
\end{aligned}
\end{equation}
Then, the gradient of $I_{\mathrm{SD3}}(s,z)$ to $\rho_{z^{c}}(s)$ becomes
\begin{equation}
    \begin{gathered}
\nabla_{\rho_{z^{c}}}I_{\mathrm{SD3}}(s,z) =\frac{\lambda d_z^\pi(s)+\rho_{z^c}(s)}{\lambda n d_z^\pi(s)}\frac{-\lambda n d_z^\pi(s)}{\left(\lambda d_z^\pi(s)+\rho_{z^c}(s)\right)^2}\\
=-\frac{1}{\lambda d_{z}^{\pi}(s)+\rho_{z^{c}}(s)}.
\end{gathered}
\end{equation}
This completes the proof.
\end{proof}

% \appendices
\section{Proof of Theorem~\ref{thm:1}}
\label{sec:proof_thm1}
\begin{proof}
For clarity, we write $I_{\rm SD3}$ as $I_{\rm SD3}(\lambda)$ to explicitly highlight its dependency on the parameter $\lambda$ in the following context.
We first note that the function $I_{\rm SD3}(\lambda)$ is monotonically increasing relative to $\lambda$, and $I_{\rm SD3}(\lambda)$ is equal to $I(S;Z)$ when $\lambda =1$. Therefore,  the first inequality
$$
I(S;Z) \leq I_{\rm SD3}(\lambda) 
$$
always holds for $\lambda \geq 1$. Next, it remains to prove the second inequality, which suffices to give an upper bound of $I_{\rm SD3}(\lambda) - I(S;Z)$. Here, for each $(s,z)$, we define $\eta_{z^c}^\pi(s) \triangleq \sum_{z' \ne z} d_{z'}^{\pi}(s)p(z')$. Note that 
\begin{equation}
\begin{aligned}
    \label{eq: upbthmt}
        & I_{\rm SD3}(\lambda) - I(S;Z) \\
        = & \EE_{z\sim p(z),s\sim d^{\pi}_{z}(s)}\left[ \log \left( \frac{\lambda\: d^{\pi}_{z}(s)}{\lambda\: d^{\pi}_{z}(s) p(z) + \eta_{z^c}^\pi(s)} \right.\right. \\
        & \quad\quad\quad\quad\quad\quad\quad \left. \left. \cdot \frac{d^{\pi}_{z}(s) p(z) + \eta_{z^c}^\pi(s)}{d^{\pi}_{z}(s)} \right) \right] \\
        = & \log\lambda + \EE_{z\sim p(z),s\sim d^{\pi}_{z}(s)}\left[ \log \frac{d^{\pi}_{z}(s) p(z) + \eta_{z^c}^\pi(s)}{\lambda\: d^{\pi}_{z}(s) p(z) + \eta_{z^c}^\pi(s)} \right] \\
        = & \log\lambda - \EE_{z\sim p(z),s\sim d^{\pi}_{z}(s)}\left[ \log \left( \frac{d^{\pi}_{z}(s) p(z) + \eta_{z^c}^\pi(s)}{d^{\pi}_{z}(s) p(z) + \eta_{z^c}^\pi(s)} \right.\right. \\
        & \quad\quad\quad\quad\quad\quad\quad \left. \left. + \frac{(\lambda - 1) d^{\pi}_{z}(s) p(z)}{d^{\pi}_{z}(s) p(z) + \eta_{z^c}^\pi(s)} \right) \right] \\
        = & \log\lambda - \EE_{z\sim p(z),s\sim d^{\pi}_{z}(s)}\left[ \log \left( 1 + (\lambda - 1) \right.\right. \\
        & \quad \quad\quad\quad\quad\quad\quad \left. \left. \cdot \frac{d^{\pi}_{z}(s) p(z)}{d^{\pi}_{z}(s) p(z) + \eta_{z^c}^\pi(s)} \right) \right].
\end{aligned}
\end{equation}

Recalling that $d^{\pi}_{z}(\cdot)$ denotes the state density of skill $z$, and $p(z)$ is the probability density function of skill $z$, we know that the term
\begin{equation}
    \log\left(1+(\lambda-1)\frac{d_z^\pi(s)p(z)}{d_z^\pi(s)p(z)+\sum_{z'\neq z}d_{z'}^\pi(s)p(z')}\right)
\end{equation}
is always non-negative for $\lambda\geq1$. Therefore, we have
\begin{equation}
    I_{SD3}(\lambda)\leq\log\lambda+I(S;Z).
\end{equation}
This completes the proof.
\end{proof}

\section{Proof of Theorem \ref{thm:2}}
\label{sec:proof_thm2}
\begin{proof}
In this proof, we first give a formulation of the intrinsic reward in a linear parameterized assumption, and then discuss the special case of tabular MDPs. 

With linear assumptions, we denote $\eta(s_t, z_t)\in\mathbb{R}^d$ as the feature vector of $(s_t,z_t)$, which is extracted by the \emph{encoder} network of CVAE. The decoder network is assumed to be a linear function of the feature vector as $\hat{s}_t=W_t \eta(s_t, z_t)$, where $W_t\in\mathbb{R}^{c\times d}$ and $\hat{s}_t\in\mathbb{R}^{c}$.
Then the reconstruction of the state becomes a regularized least-squared problem that captures the prediction error given a dataset $\mathcal{D}_m$, where $m$ is the number of episodes in the dataset. Thus,  we have
\begin{equation}
\label{app: linear least-square}
W_t = \arg\min_{W}\sum_{i=0}^{m}\bigl\|s^i_{t}-W\eta(s^i_t,z^i_t)\bigr\|^2_F + \kappa\cdot \|W\|^2_F,
\end{equation}
where $\|\cdot\|_F$ denotes the Frobenius norm. 
% Intuitively, it can be seen as the linear case in our CVAE objective, 
We further define the following noise with respect to the least-square problem in Eq.~\eqref{app: linear least-square} as
\begin{equation}
    \begin{aligned} 
    \label{app: as-s}
        s_t = W_t\eta(s_t,z_t) + \epsilon, \quad \epsilon \sim \mathcal{N}(0,\mathbf{I}).
    \end{aligned}
\end{equation}
% \mathcal{N}(0, \mathbf{I})
Here we consider the estimation error $\epsilon$ in Eq.~\eqref{app: linear least-square} to follow the standard multivariate Gaussian distribution.

Recall that our practical intra-skill exploration reward is $D_{\rm KL}[Q_\phi(h|s,z)\| ~r(h)]$, where $Q_\phi$ is a posterior network compressing the representation of each state and skill with parameter $\phi$, and $r(h)$ is 
% With a slight abuse of notation, we denote $Q^{\rm margin}$ as 
the marginal distribution of the latent variable,
% of the encoder over the posterior of the parameter $\phi$. 
% Note that the posterior $Q^{\rm margin}$ is intractable, 
where we follow previous works \cite{alemi2016deep, bai2021dynamic} to consider the marginal as the standard normal distribution. 
% With linear assumption, we also consider the encoder $Q_\phi(h|s,z)$ is a linear mapping of $(s,z)$ and denoted as $h=Q^H(h|s,z;\phi)$.
% Building on the groundwork laid out above, the intra-skill exploration reward is equivalent to the KL divergence of $Q^{\rm margin}$ from $Q_\phi(h|s,z)$. In other words, with $Q^{\rm margin}$ is approximated as $\mathcal{N}(0, \mathbf{I})$, we have
% \begin{equation}
%     \begin{aligned}
%     r^{\rm exp}_{z_t}(s_t) = D_{\rm KL}\left[Q_\phi(\cdot|s_t,z)\| ~Q^{\rm margin}\right].
%     \end{aligned}
% \end{equation}
Then we re-define the intrinsic reward in a Bayesian perspective, where we introduce $\Phi$ to denote the total parameters, as
\begin{equation}
\begin{aligned}
r^{\rm exp}_z(s) &= \EE_{\Phi} D_{\rm KL}[Q_\phi(h|s,z)\| ~r(h)]\\
&=\cH(Q^{\rm margin}) - \cH(Q_\phi(h|s,z)),
\end{aligned}
\end{equation}
where $Q^{\rm margin}=Q(s,z)|\cD_m$ is the margin distribution of the encoding over the posterior of the parameters $\Phi$. In practice, we replace the expectation over posterior $\Phi$ by the corresponding point estimation, namely the parameter $\phi$ of the neural networks trained with SD3 model on the dataset $\cD_m$.
Formally, considering the Bayesian form of learning objective, we have
\begin{equation}
\begin{aligned}
r^{\rm exp}_z(s)&=\cH(Q^{\rm margin}) - \cH(Q_\phi(h|s,z))\\
&=\cH(Q(s,z,S)| \cD_m) - \cH(Q(s,z,S)| \Phi,\cD_m),
\end{aligned}
\end{equation}
where $Q$ is a neural network in practice. We adopt the mapping 
\begin{equation}
Q(s,z,S)| \Phi,\cD_m=Q_\phi(h|s,z)
\end{equation}
since $Q_\phi$ is trained to reconstruct the variable $S$, where $\phi$ constitutes a part of the parameters of the total parameters $\Phi$. According to Data Processing Inequality, the post-processing of the signal does not increase information, and we can understand $Q$ as post-processing mapping the state-skill vector via an encoder network. Then we have the following inequality for the information-gain term:
\begin{equation}
\label{eq:app-exp-inq}
\begin{aligned}
r^{\rm exp}_z(s)&=\cH(Q(s,z,S)| \cD_m) - \cH(Q(s,z,S)| \Phi,\cD_m) \\
&\leq \cH(s,z,S| \cD_m) - \cH(s,z,S| \Phi,\cD_m)\\
&=I(\Phi;(s,z,S)|\cD_m),
\end{aligned}
\end{equation}
where we denote $(s,z)$ as realizations as they are sampled from the dataset as input, and $S$ is a random variable that is learned to reconstruct by parameter $\Phi$. The inequality can be tight since $Q(\cdot)$ is trained by reconstruction, which contains sufficient information about $(s,z,S)$. 

In the following, we will prove the following inequality in a linear case with a parameter $W_t$ considered in Eq.~\eqref{app: linear least-square}, as
\begin{equation}
\begin{aligned}
\label{eq:remain}
r^{\rm exp}_{z_t}(s_t)&\leq I(W_t;(s_t,z_t,S_t)|\cD_m)\\
& \leq \frac{c}{2} [ \eta(s_t,z_t)^{\top}\Lambda_t^{-1}\eta(s_t,z_t)],
\end{aligned}
\end{equation}
% where $\beta_0$ is a tuning hyper-parameter, 
the $[ \eta(s_t,z_t)^{\top}\Lambda_t^{-1}\eta(s_t,z_t)]$ term is known as an upper-confidence-bound (UCB)-term in linear MDPs \cite{lsvi-ucb,cai2020provably}, and $\Lambda_t=\sum_{j=1}^m \eta(s_j,z_j) \eta(s_j,z_j)^\top + \kappa \cdot \mathbf{I}$ is the covariance matrix of the samples in the dataset. Finally, we will connect the UCB-term to the count-based bonus in the tabular case. 

Let denote ${\rm vec}(W_t)$ as vectorization of $W_t\in\mathbb{R}^{c\times d}$,
\begin{equation}\label{eq::vecw}
{\rm vec}(W_t)=
\begin{bmatrix}
\begin{smallmatrix}
w_{11} & \cdots & w_{1d} &w_{21} &\cdots &w_{2d} &\cdots &\cdots &w_{c1} &\cdots &w_{cd}
\end{smallmatrix}
\end{bmatrix}^{\top}\in\mathbb{R}^{cd},
\end{equation}
and also $\tilde\eta(s_t,z_t)$, 
\begin{equation}
\label{eq::eta}
\begin{aligned}
\tilde{\eta}(s_t,z_t) &=
\begin{bmatrix}
\begin{smallmatrix}
\eta(s_t,z_t) & 0             & \cdots & 0            \\
0             & \eta(s_t,z_t) & \cdots & 0            \\
\vdots        & \vdots        & \ddots & \vdots       \\
0             & 0             & \cdots & \eta(s_t,z_t)\\
\end{smallmatrix}
\end{bmatrix}\\
&=
\begin{bmatrix}
\begin{smallmatrix}
\eta_{1} & \cdots & \eta_{d} & & \cdots & & \cdots &\cdots   & 0    & \cdots & 0  \\
0     &   & 0       & \eta_{1} & \cdots & \eta_{d} &\cdots & \cdots & 0 &\cdots &0       \\
\vdots   && \vdots  & \vdots  & & \vdots  &        && \vdots  &&\vdots\\
0       && 0  &0  & \cdots & 0   & \cdots &\cdots & \eta_{1} & \cdots & \eta_{d} \\
\end{smallmatrix}
\end{bmatrix}^{\top}
\in \mathbb{R}^{cd\times c},
\end{aligned}
\end{equation}
then it is not difficult to verify that ${\rm vec}(W_t)^{\top}\tilde{\eta}(s_t,z_t)=W_t\eta(s_t,z_t)$. By the definition of the mutual information, we observe 
\begin{equation}
\label{eq:vecw-info}
\begin{aligned}
&\quad I(W_t;[s_t,z_t,S_{t}]\mid \mathcal{D}_m)\\
&= I({\rm vec}(W_t);[s_t,z_t, S_{t}]\mid \mathcal{D}_m)\\&=\mathcal{H}({\rm vec}(W_t)\mid \mathcal{D}_m)-\mathcal{H}({\rm vec}(W_t)\mid \mathcal{D}_m \cup(s_t,z_t, S_{t}))\\
&=\frac{1}{2}\log\det\big({\rm Var}({\rm vec}(W_t)\mid \mathcal{D}_m)\big)\\
&-\frac{1}{2}\log \det\big({\rm Var}({\rm vec}(W_t)\mid \mathcal{D}_m \cup(s_t,z_t, S_{t}))\big).
\end{aligned}
\end{equation}
Next, we need to obtain ${\rm Var}({\rm vec}(W_t)\mid \mathcal{D}_m)$ and ${\rm Var}({\rm vec}(W_t)\mid \mathcal{D}_m \cup(s_t,z_t, S_{t}))$. Recalling that $\epsilon$ satisfies the standard Gaussian distribution in Eq.~\eqref{app: as-s}, we can conclude that 
$$s_t | \eta_t,W_t \sim \mathcal{N}({\rm vec}(W_t)^{\top} \tilde{\eta}(s_t, z_t), \mathbf{I}).$$ 

Assuming the prior distribution $W\sim \cN(0,\mathbf{I}/\kappa) $, then the prior of ${\rm vec}(W)$ also follows from $\cN(0,\mathbf{I}/\kappa)$. Moreover, using Bayes' theorem and plugging the probability of $p({\rm vec}(W_t))$, we have 
\begin{equation}
\label{eq:pf_bayes_rule}
\begin{aligned}
&\quad \log p({\rm vec}(W_t) \mid \mathcal{D}_m)\\
&= \log p({\rm vec}(W_t)) + \log p(\mathcal{D}_m \mid {\rm vec}(W_t)) - \log p(\mathcal{D}_m) 
\\&=- \|{\rm vec}(W_t)\|^2/2 -\sum^m_{i=1} \| {\rm vec}(W_t)\tilde{\eta}(s^i_t, z^i_t) - s^i_{t+1}\|^2/2 + \text{C}
\\&=-({\rm vec}(W_t)-\tilde{\mu}_{t,m})^\top \tilde{\Lambda}^{-1}_{t,m}({\rm vec}(W_t)-\tilde{\mu}_{t,m})/2 + \text{C},
\end{aligned}
\end{equation}
where $\tilde{\mu}_t$ and $\tilde{\Lambda}_t$ in the last equality are defined as 
\begin{equation}
\begin{aligned}
&\tilde{\mu}_{t,m} = \tilde{\Lambda}^{-1}_t \sum^m_{i=0}\tilde{\eta}(s^i_t, z^i_t) s^i_{t+1}\in\mathbb{R}^{cd},\\
&\tilde{\Lambda}_{t,m}=\sum_{i=0}^{m}\tilde{\eta}(s_t^i,z_t^i)\tilde{\eta}(x_t^i,z_t^i)^\top+\kappa \cdot \mathrm{\mathbf{I}}\in\mathbb{R}^{cd\times cd}.
\end{aligned}
\end{equation}
Taking the left-hand side of $\log$ to the right. The Eq.~\eqref{eq:pf_bayes_rule} implies the distribution of ${\rm vec}(W_t)\mid \mathcal{D}_m \sim  N(\tilde{\mu}_{t,m}, \tilde{\Lambda}^{-1}_{t,m})$. Hence, we can get
\begin{equation}
\label{eq:var-vec-W}
    \begin{aligned}
    &{\rm Var}({\rm vec}(W_t)\mid \mathcal{D}_m) = \tilde{\Lambda}^{-1}_{t,m},\\ 
    &{\rm Var}({\rm vec}(W_t)\mid \mathcal{D}_m \cup(s_t,z_t, S_{t})) = \tilde{\Lambda}^{-1}_{t,m+1}.
    \end{aligned}
\end{equation}
We proceed to derive Eq.~\eqref{eq:vecw-info} by applying Eq.~\eqref{eq:var-vec-W}, from which we obtain
\begin{equation}
\label{eq:vecw-info1}
\begin{aligned}
&\quad I({\rm vec}(W_t);[s_t,z_t, S_{t+1}]|\mathcal{D}_m)\\
&=\frac{1}{2}\log\det\big(\tilde{\Lambda}^{-1}_{t,m}\big)-\frac{1}{2}\log \det\big(\tilde{\Lambda}^{-1}_{t,m+1}\big)\\
&=\frac{1}{2}\log\det\big(\tilde{\Lambda}_{t,m+1}+\tilde{\eta}(s_t,z_t)\tilde{\eta}(s_t,z_t)^{\top}\big)-\frac{1}{2}\log \det\big(\tilde{\Lambda}_{t,m}\big)\\
&=\frac{1}{2}\log\det\big(\tilde{\eta}(s_t,z_t)^{\top}\tilde{\Lambda}_{t}^{-1}\tilde{\eta}(s_t,z_t)+\mathbf{I}\big),
\end{aligned}
\end{equation}
where the last equality holds by applying the Matrix Determinant Lemma to the first term. Recalling our definition of $\tilde{\eta}(s_t,z_t)$, the state-skill pairs are finite in the tabular case, so we have 
\begin{equation}
\begin{aligned}
\tilde{\Lambda}_t&=\sum_{i=0}^{m}\tilde{\eta}(s_t^i,z_t^i)\tilde{\eta}(x_t^i,z_t^i)^\top+\kappa\cdot  \mathrm{\mathbf{I}}	\\ &=
\begin{bmatrix}
\begin{smallmatrix}
a_0+\kappa I & 0 & \cdots & 0     \\
 0     & a_1+\kappa I & \cdots & 0     \\
\vdots & \vdots & \ddots & \vdots\\
0      & 0 & \cdots & a_m+\kappa I\\
\end{smallmatrix}
\end{bmatrix},
a_k=\sum\eta(s_k,z_k)\eta(s_k,z_k)^{\top}.
\end{aligned}
\end{equation}
Then, $\tilde{\eta}(s_t,z_t)^{\top}\tilde{\Lambda}_{t}^{-1}\tilde{\eta}(s_t,z_t)$ can be rewritten as 
\begin{equation}
\label{app:resdapm}
\begin{split}
\tilde{\eta}&(s_t,z_t)^{\top}\tilde{\Lambda}_{t}^{-1}\tilde{\eta}(s_t,z_t)\\ =&
% \begin{bmatrix}
% \begin{smallmatrix}
% \eta(s_t,z_t)^{\top} & 0      & \cdots & 0     \\
%  0     & \eta(s_t,z_t)^{\top} & \cdots & 0         \\
% \vdots & \vdots         & \ddots & \vdots\\
% 0      & 0              & \cdots & \eta(s_t,z_t)^{\top}
% \end{smallmatrix}
% \end{bmatrix}
% \begin{bmatrix}
% \begin{smallmatrix}
% \Lambda^{-1} & 0 & \cdots & 0     \\
%  0        & \Lambda^{-1} & \cdots & 0     \\
% \vdots    & \vdots & \ddots & \vdots\\
% 0         & 0 & \cdots & \Lambda^{-1} 
% \end{smallmatrix}
% \end{bmatrix}
% \begin{bmatrix}
% \begin{smallmatrix}
% \eta(s_t,z_t) & 0      & \cdots & 0     \\
%  0     & \eta(s_t,z_t) & \cdots & 0         \\
% \vdots & \vdots         & \ddots & \vdots\\
% 0      & 0              & \cdots & \eta(s_t,z_t)
% \end{smallmatrix}
% \end{bmatrix}
% \\ = &
\begin{bmatrix}
% \begin{smallmatrix}
\xi & 0      & \cdots & 0     \\
 0     & \xi & \cdots & 0         \\
\vdots & \vdots         & \ddots & \vdots\\
0      & 0       & \cdots        & \xi
% \end{smallmatrix}
\end{bmatrix}\in\mathbb{R}^{c\times c},
\quad \xi = \eta(s_t,z_t)^{\top}\Lambda^{-1}\eta(s_t,z_t).
\end{split}
\end{equation}
Therefore, by eliminating the determinant based on the expression in Eq.~\eqref{app:resdapm} and applying the inequality $\log(1 + x) \leq x$ for $x\geq 0$, we can further bound Eq~\eqref{eq:vecw-info1} from above as
\begin{equation}
    \begin{aligned}
    \label{app: ineq-info-count}
        &\quad I({\rm vec}(W_t);[s_t,z_t, S_{t+1}]|\mathcal{D}_m)\\
        &=\frac{1}{2}\cdot\log\det\left(\tilde{\eta}(s_t,z_t)^{\top}\tilde{\Lambda}_{t}^{-1}\tilde{\eta}(s_t,z_t)+\mathbf{I}\right) \\&= \frac{c}{2}\cdot\log\left(\eta(s_t,z_t)^{\top}\Lambda^{-1}\eta(s_t,z_t)+ 1\right) \\ &\leq \frac{c}{2}\cdot\eta(s_t,z_t)^{\top}\Lambda^{-1}\eta(s_t,z_t).
    \end{aligned}
\end{equation}
Hence, based on Eq.~\eqref{eq:app-exp-inq} and Eq.~\eqref{app: ineq-info-count}, we conclude that
\begin{equation}
\begin{aligned}
r^{\rm exp}_{z}(s_t) &\leq I(W_t;[s_t,z_t, S_{t+1}]|\mathcal{D}_m)\\ 
&= I({\rm vec}(W_t);[s_t,z_t, S_{t+1}]|\mathcal{D}_m)\\
&\leq \frac{c}{2}\cdot\eta(s_t,z_t)^{\top}\Lambda^{-1}\eta(s_t,z_t).
\end{aligned}
\end{equation}

In tabular cases \cite{auer2006logarithmic}, the state and skill are considered as finite and countable. Let $d= |\cS| \times |\cZ|$. Recall that $\eta(s_t,z_t) \in \mathcal{R}^{|\cS||\cZ|}$ is the one-hot vector with a value of 1 at position $(s_t, z_t) \in \cS \times \cZ$, i.e.,
\begin{equation}
\begin{aligned}
&\eta(s_j,z_j)=
\begin{bmatrix}
0\\
\vdots\\
1\\
\vdots\\
0
\end{bmatrix}
\in \mathbb{R}^d,
\quad \text{and}\\
&\eta(s_j,z_j)\eta(s_j,z_j)^{\top}=
\begin{bmatrix}
0      & \cdots & 0 & \cdots & 0\\
\vdots & \ddots &   &        & \vdots \\
0      &        & 1 &        & 0\\
\vdots &        &   & \ddots & \vdots \\
0      & \cdots & 0 & \cdots & 0\\
\end{bmatrix}
\in \mathbb{R}^{d\times d}.
\end{aligned}
\end{equation}
We denote the gram matrix $\Lambda_j =\sum_{i=0}^{m}\eta(s_j^i,z_j^i)\eta(s_j^{i},z_j^{i})^\top+\kappa \cdot \mathrm{\mathbf{I}}$ for $\kappa > 0$ as covariance matrix given a dataset $\mathcal{D}_m$. Since we denote $\eta$ as a one-hot vector, and $\Lambda$ as the sum of all the matrices $\eta(s_j,z_j)\eta(s_j,z_j)^{\top}$, each diagonal element of $\Lambda$ can be seen as the corresponding count $N(s_j ,z_j)$ for the state-skill pair, i.e.
\begin{align*}
\small
\Lambda=
\begin{bmatrix}
a_0+\kappa & 0 & \cdots & 0     \\
 0     & a_1+\kappa & \cdots & 0     \\
\vdots & \vdots & \ddots & \vdots\\
0      & 0 & \cdots & a_m+\kappa\\
\end{bmatrix},
\quad a_k = N(s_k ,z_k).
\end{align*}
Moreover, given a dataset, the expression on the right side of the theorem's inequality is inversely proportional to the total number of state-skill pairs; in other words, 
\begin{equation}
\begin{aligned}
\label{app: inver-porpotional count}
\eta(s_j,z_j)^{\top}\Lambda_t^{-1}\eta(s_j,z_j)=\frac{1}{N(s_j ,z_j)+\kappa}.
\end{aligned}
\end{equation}

According to Eq.~\eqref{eq:remain}, we have the following relationship in the tabular case:
\begin{equation}
\begin{aligned}
r^{\rm exp}_{z_t}(s_t)&\leq I(W_t;(s_t,z_t,S_t)|\cD_m)\\
&\leq \frac{c}{2} [ \eta(s_t,z_t)^{\top}\Lambda_t^{-1}\eta(s_t,z_t)] = \frac{|\mathcal{S}|/2}{N(s_t ,z_t)+\kappa}.
\end{aligned}
\end{equation}
The first inequality is due to the Data Processing Inequality according to Eq.~\eqref{eq:app-exp-inq}. The bound is tight since $Q(\cdot)$ is trained by reconstruction, which contains sufficient information about $(s,z,S)$. The second inequality is tight when $ \eta(s_t,z_t)^{\top}\Lambda_t^{-1}\eta(s_t,z_t)\rightarrow 0$, which means that the count of state-action pair is large. In the last equation, $c$ is the count of all states in the tabular space. Thus, we have
\begin{equation}
r^{\rm exp}_{z}(s) \approx \frac{|\mathcal{S}|/2}{N(s ,z)+\kappa},
\end{equation}
if the count of $N(s ,z)$ is large. Intuitively, optimizing the reward $\eta(s,z)^{\top}\Lambda^{-1}\eta(s,z)$ incentivizes the agent to increase the visitation of $(s, z)$. Furthermore, since we have proven that Eq.~\eqref{eq:remain} holds, we can state that in the tabular case, maximizing the intra-skill reward is equivalent to maximizing the count-based rewards \cite{bellemare2016unifying,ostrovski2017count}. The intra-skill exploration reward encourages the skill-conditional policy to increase the visitation times of those rare state-skill pairs.

\end{proof}

\bibliography{main}
\bibliographystyle{IEEEtran}

\vfill

\end{document}